\documentclass{article}
\usepackage{amsmath}
\usepackage{graphicx}
\usepackage{amsfonts}
\usepackage{amssymb}
\newtheorem{theorem}{Theorem}

\newtheorem{corollary}[theorem]{Corollary}

\newtheorem{definition}[theorem]{Definition}
\newtheorem{example}[theorem]{Example}

\newtheorem{lemma}[theorem]{Lemma}

\newtheorem{remark}[theorem]{Remark}

\begin{document}

\title{Concentration inequalities of the cross-validation estimate for stable predictors}
\author{Matthieu Cornec}
\maketitle
\begin{abstract}
\bigskip

In this article, we derive concentration inequalities for the cross-validation
estimate of the generalization error for stable predictors in the context of
risk assessment. The notion of stability has been first introduced by
\cite{DEWA79} and extended by \cite{KEA95} , \cite{BE01} and \cite{KUNIY02} to
characterize class of predictors with infinite VC dimension. In particular,
this covers $k$-nearest neighbors rules, bayesian algorithm (\cite{KEA95}),
boosting,$\ldots$ General loss functions and class of predictors are
considered. We use the formalism introduced by \cite{DUD03} to cover a large
variety of cross-validation procedures including leave-one-out
cross-validation, $k$-fold cross-validation, hold-out cross-validation (or
split sample), and the leave-$\upsilon$-out cross-validation. \bigskip

\noindent In particular, we give a simple rule on how to choose
the cross-validation, depending on the stability of the class of
predictors. In the special case of uniform stability, an
interesting consequence is that the number of elements in the
test set is not required to grow to infinity for the consistency
of the cross-validation procedure. In this special case, the
particular interest of leave-one-out cross-validation is
emphasized.

\bigskip

\end{abstract}

%
%
%

%
%
%

\addcontentsline{toc}{section}{Introduction} \markboth{\uppercase
{Introduction}} {\uppercase{Introduction}}

\bigskip

\newpage

\section{Introduction and motivation}

\noindent One of the main issue of pattern recognition is to
create a predictor (a regressor or a classifier) which takes
observable inputs in order to predict the unknown nature of an
output. Formally, a predictor $\varphi$ is a measurable map from
some measurable space $\mathcal{X}$ to some measurable space
$\mathcal{Y}$. When $\mathcal{Y}$ is a countable set (respectively
$\mathbb{R}^{m}$), the predictor is called a classifier
(respectively a regressor). The strategy of \textit{Machine
Learning }consists in building a learning algorithm $\Phi$ from
both a set of examples and a class of methods. Typical class of
methods are empirical risk minimization or $k$-nearest neighbors
rules. The set of examples consists in the measurement of $n$
observations $(x_{i},y_{i})_{1\leq i\leq n}$. Thus, formally,
$\Phi$ is a
measurable map from $\mathcal{X}\times\mathcal{\cup}_{n}(\mathcal{X}%
\times\mathcal{Y)}^{n}$ to $\mathcal{Y}$. One of the main issue of
\textit{Statistical Learning }is to analyze the performance of a
learning algorithm in a probabilistic setting.
$(x_{i},y_{i})_{1\leq i\leq n}$ are supposed to be observations
from $n$ independent and identically distributed (i.i.d.) random
variables $(X_{i},Y_{i})_{1\leq i\leq n}\ $with unknown
distribution $\mathbb{P}$. $(X_{i},Y_{i})_{1\leq i\leq n}$ is
denoted $\mathcal{D}_{n}$ in the following and called the
learning set. In order to analyze the performance, it is usual to
consider the conditional risk of a machine learning $\Phi$
denoted $\tilde{R}_{n}$, so called the generalization error. It is
defined by the conditional expectation of
$L(Y,\Phi(X,\mathcal{D}_{n}))$ given $\mathcal{D}_{n}$ where
$(X,Y)\sim\mathbb{P}$ is a random variable
independent of $\mathcal{D}_{n}$, i.e. $\tilde{R}_{n}:=\mathbb{E}%
_{X,Y}(L(Y,\Phi(X,\mathcal{D}_{n}))|\mathcal{D}_{n})$ with $L$ a
cost function from
$\mathcal{Y}^{2}\longrightarrow\mathbb{R}_{+}$. Notice that
$\widetilde{R}_{n}$ is a random variable measurable with respect
to $\mathcal{D}_{n}$.

\bigskip

\noindent An important question is: the distribution $\mathbb{P}$ of the
generating process being unknown, can we estimate how good a predictor trained
on a learning set of size $n$ is? In other words, can we approximate the
generalization error $\widetilde{R}_{n}$? This fundamental statistical problem
is referred to ''choice and assessment of statistical
predictions''\ \cite{STO74}. Many estimates have been proposed.
%
%
%
Quoting \cite{HTF01}: \textit{Probably the simplest and most widely used
method for estimating prediction error is cross-validation}.

\bigskip

\noindent The cross-validation procedures include leave-one-out
cross-validation, $k$-fold cross-validation, hold-out cross validation (or
split sample), leave-$\upsilon$-out cross-validation (or Monte Carlo
cross-validation or bootstrap cross-validation). With the exception of
\cite{BUR89}, theoretical investigations of multifold cross-validation
procedures have first concentrated on linear models (\cite{Li87}%
;\cite{SHAO93};\cite{ZHA93}). Results of \cite{DGL96} and \cite{GYO02} are
discussed in Section 3. The first finite sample results are due to Wagner and
Devroye \cite{DEWA79} and concern $k$-local rules algorithms under
leave-one-out and hold-out cross-validation. More recently, \cite{HOL96,
HOL96bis} derived finite sample results for $\upsilon$-out cross-validation,
$k-$fold cross-validation, and leave-one-out cross-validation for Empirical
Risk Minimization (ERM) over a class of predictors with finite
Vapnik-Chervonenkis-dimension (VC-dimension)in the realisable case (the
generalization error is equal to zero). \cite{BKL99} have emphasized when
$k-$fold can beat $\upsilon$-out cross-validation in the particular case of
$k$-fold predictor. \cite{KR99} has extended such results in the case of
stable algorithms for the leave-one-out cross-validation procedure.
\cite{KEA95} also derived results for hold-out cross-validation for ERM, but
their arguments rely on the traditional notion of VC-dimension. In the
particular case of ERM over a class of predictors with finite VC-dimension but
with general cross-validation procedures, we derived derived probability
upper bounds in chapter 1: we denote by $p_{n}$ the percentage of elements in the test
sample. In the sequel, we will denote by $\widehat{R}_{CV}$ the
cross-validation estimator. For empirical risk minimizers over a class of
predictors with finite VC-dimension $V_{\mathcal{C}}$, to be defined below,
we obtained the following concentration inequality. For all
$\varepsilon>0,$ we have%
\[
\Pr\mathbb{(}|\widehat{R}_{CV}-\widetilde{R}_{n}|\geq\varepsilon)\leq
B(n,p_{n},\varepsilon)+V(n,p_{n},\varepsilon),
\]

with

\begin{itemize}
\item $B(n,p_{n},\varepsilon)=\displaystyle5(2n(1-p_{n})+1)^{\frac
{4V_{\mathcal{C}}}{1-p_{n}}}\exp(-\frac{n\varepsilon^{2}}{64}),$

\item $V(n,p_{n},\varepsilon)=\displaystyle\min\left(  \exp(-\frac
{2np_{n}\varepsilon^{2}}{25}),\frac{16}{\varepsilon}\sqrt{\frac{V_{\mathcal{C}%
}(\ln(2(1-p_{n})+1)+4)}{n(1-p_{n})}}\right)  .$
\end{itemize}

\noindent Unfortunately, many popular predictors, including
$k$-nearest neighbors rules, do not satisfy this property.
Moreover, these bounds obtained are called ''sanity check
bounds''\ since they are not better than classical
Vapnik-Chernovenkis's bounds.

\bigskip

\noindent To avoid the traditional analysis in the VC framework, notions of
stability have been intensively worked through in the late 90's \cite{KEA95},
\cite{BE01}, \cite{BE02}, \cite{KUT02}, and \cite{KUNIY02}. The object of
stability framework is the learning algorithm rather than the space of
classifiers. The learning algorithm is a map (effective procedure) from data
sets to classifiers. An algorithm is stable at a learning set $\mathcal{D}%
_{n}$ if changing one point in $\mathcal{D}_{n}$ yields only a small change in
the output hypothesis. The attraction of such an approach is that it avoids
the traditional notion of VC-dimension, and allows to focus on a wider class
of learning algorithms than empirical risk minimization. For example, this
approach provides generalization error bounds for regularization-based
learning algorithms that have been difficult to analyze within the VC
framework such as boosting. As a motivation, we quote the following list of
algorithms satisfying stability properties: regularization networks, ERM,
k-nearest rules, boosting.

\bigskip

\noindent Algorithmic stability was first introduced by \cite{DEWA79}.
\cite{BRE96} argued that unstable weak learners benefit from randomization
algorithms such as bagging. \cite{KR99} considered both algorithmic stability
and the weaker related notion of error stability. They proved bounds on the
error of cross-validation estimates of generalization error, but their
arguments rely on VC theory. \cite{BE01,BE02} proved that an algorithm which
is stable everywhere has low generalization error; their proof does not make
any reference to VC-dimension. They showed that regularization networks are
stable. In \cite{KUNIY02}, at least ten different notions were examined. In
particular, they introduced a probabilistic notion of change-one stability
called Cross-Validation stability or CV stability. This was shown to be
necessary and sufficient for consistency of ERM in the Probably Approximately
Correct (PAC) Model of \cite{VAL84}.

\bigskip

\noindent The goal of this paper is to obtain exponential bounds to fill the
chart \ref{Chart} where possible bounds are missing (up to our knowledge).

{\tiny \begin{table}[th]
\begin{center}
{\tiny {\scriptsize
\begin{tabular}
[c]{rrrrr}\hline
& leave-one-out & hold-out & k-fold & $\nu$-out\\\hline\hline
ERM with \\ finite VC-dimension & Kearns, Holden, Cornec &
Holden,Cornec & Holden,Cornec & Cornec\\
hypothesis stability & Devroye and W & Devroye and W & $\times$ & $\times$\\
error stability \\ with finite VC dimension & Kearns & Kearns & $\times$ &
$\times$\\
uniform stability & Bousquet and E. & $\times$ & $\times$ & $\times$\\
strong hypothesis & Kutin and N & $\times$ & $\times$ & $\times$\\
weak stability & $\times$ & $\times$ & $\times$ & $\times$\\\hline
\end{tabular}
}  }
\end{center}
\caption{Missing bounds $\times$ to find}%
\end{table}\label{Chart}}

\bigskip

\noindent The goal of this article is also to show that cross-validation is
still consistent for stable predictors. As a consequence, we will emphasize
the role played by cross-validation: it can be a consistent estimate of the
generalisation error when the training error defined by $\widehat{R}%
_{n}:=\frac{1}{n}\sum_{i=1}^{n}L(Y_{i},\phi(X_{i},\mathcal{D}_{n}))$ is not.
Indeed, for stable predictors, the training error can be arbitrarly poor: for
example, the training error for $1$-nearest neighboor is equal to zero
whatever the generalisation error may be.
%
%
%

\bigskip

\noindent We introduce our \textbf{main result}\footnote{accurate inequalities
can be found in section \ref{results}}. Suppose that the cross-validation is
symmetric --i.e. the probability of a observation to be in the training set is
independent of its index- and that the number of elements in the test set is
constant and equal to $np_{n}$ with $p_{n}$ the percentage of elements in the
test set. All the bounds of the following form $\Pr\mathbb{(}|\widehat{R}%
_{CV}-\widetilde{R}_{n}|\geq\varepsilon+\ldots)\leq
B(n,p_{n},\varepsilon )+V(n,p_{n},\varepsilon)$.

\noindent Under certain stability conditions -satisfied for example by
Empirical Risk Minimisers (ERM) or Adaboost-, we have for all $\varepsilon
\geq0$,
\[
\Pr\mathbb{(}|\widehat{R}_{CV}-\widetilde{R}_{n}|\geq\varepsilon+2\lambda
p_{n})\leq2\exp(-2np_{n}\varepsilon^{2})+\delta_{n,p_{n}}%
\]
\noindent with $\delta_{n,p_{n}}$ and $\lambda$ a non-negative
real numbers.
For classical algorithms, we have in mind that $\delta_{n,p_{n}}=O_{n}%
(p_{n}\exp(-n(1-p_{n}))$. $\lambda$ is in fact a Lipschitz
coefficient with respect to the total variation and can be
interpreted as a stability factor: the smaller $\lambda$ is, the
more stable the learning algorithm is. Furthermore, if the
learning algorithm satisfies a stronger stability condition
(for example Adaboost or regularization networks), we obtain%
\[
\Pr\mathbb{(}|\widehat{R}_{CV}-\widetilde{R}_{n}|\geq\varepsilon
+\delta_{n,p_{n}}+2\lambda p_{n})\leq4(\exp(-\frac{\varepsilon^{2}%
}{8(18\lambda)^{2}np_{n}^{2}})+\frac{n}{9\lambda p_{n}}\delta_{n,p_{n}%
}^{^{\prime}}))
\]
\noindent with $\delta_{n,p_{n}}^{^{\prime}}=\delta_{n,p_{n}}+(n+1)\delta
_{n,1/n}$. For the latter, it is thus not required that the number of elements
in the test set grows to infinity for the consistency of the cross-validation
to hold.

\bigskip

\noindent Using these probability bounds, we can then deduce that the
expectation between the generalization error and the cross-validation error
$\mathbb{E}_{\mathcal{D}_{n}}|\widehat{R}_{CV}-\widetilde{R}_{n}|$ is of order
$O_{n}((\lambda/n)^{1/3})$. As far as the expectation $\mathbb{E}%
_{\mathcal{D}_{n}}|\widehat{R}_{CV}-\widetilde{R}_{n}|$ is concerned, we can
define a splitting rule in the general setting: the percentage of elements
$p_{n}^{\star}$ in the test set should be proportional to $(1/\lambda^{^{2}%
}n)^{1/3}$, i.e. the less stable (i.e. $\lambda$ large) the
learning algorithm is, the smaller the test set in the
cross-validation should be. Furthermore, if the learning algorithm
satisfies a stronger stability condition (for example
Adaboost or regularization networks), we also have $\mathbb{E}_{\mathcal{D}%
_{n}}|\widehat{R}_{CV}-\widetilde{R}_{n}|=O_{n}(\lambda/\sqrt{n})$ and the
leave-one-out cross-validation (i.e. $p_{n}^{\star}=1/n$) is preferred for $n$
large enough.

\bigskip

\noindent The paper is organized as follows. In the next section, we recall
the main notations and definitions of cross-validation as introduced in
chapter 1. We also introduce notations to unify the main notions of
stability. Finally, in Section 3, we introduce our results in terms of
probability upperbounds. We also prove that many traditionnal methods satisfy
our generalized notion of stability (lasso,...,adaboost, k-nearest neighbors).

\section{Notations and definitions}

\noindent In the following, we follow the notations of cross-validation
introduced in chapter 1.

\subsection{Cross-validation}

\noindent We will consider the following shorter notations inspired by the
literature on empirical processes. In the sequel, we will denote
$\mathcal{Z}:=\mathcal{X\times Y}$, and $(Z_{i})_{1\leq i\leq n}%
:=((X_{i},Y_{i}))_{1\leq i\leq n}$ \ the learning set. For a given loss
function $L$ and a given class of predictors $\mathcal{G}$, we define a new
class $\mathcal{F}$ of functions from $\mathcal{Z}$ to $\mathbb{R}_{+}$ by
$\mathcal{F}:=\{\psi\in\mathbb{R}_{+}^{\mathcal{Z}}|\psi(Z)=L(Y,\phi
(X)),\phi\in\mathcal{G}\}$. For a machine learning $\Phi$, we have the natural
definition $\Psi(Z,\mathcal{D}_{n}):=L(Y,\Phi(X,\mathcal{D}_{n})).$ With these
notations, the conditional risk $\widetilde{R}_{n}\ $is the expectation of
$\Psi(Z,\mathcal{D}_{n})$ with respect to $\mathbb{P}$ conditionally on
$\mathcal{D}_{n}$: $\widetilde{R}_{n}:=\mathbb{E}_{Z}[\Psi(Z,\mathcal{D}%
_{n})\mid\mathcal{D}_{n}]$ with $Z\sim\mathbb{P}$ independent of
$\mathcal{D}_{n}$. In the following, if there is no ambiguity, we will also
allow the following notation $\psi(X,\mathcal{D}_{n})$ instead of
$\Psi(X,\mathcal{D}_{n})$.

\noindent To define the accurate type of cross-validation procedure, we
introduce binary vectors. Let $V_{n}=(V_{n,i})_{1\leq i\leq n}$ be a vector of
size $n$. $V_{n}$ is a binary vector if for all $1\leq i\leq n,V_{n,i}%
\in\{0,1\}$ and if $\sum_{i=1}^{n}V_{n,i}\neq0$. Consequently, we can define
the subsample associated with it, $\mathcal{D}_{V_{n}}:=\{Z_{i}\in
\mathcal{D}_{n}|V_{n,i}=1,1\leq i\leq n\}$. We define a weighted empirical
measure on $\mathcal{Z}$%
\[
\mathbb{P}_{n,V_{n}}:=\frac{1}{\sum_{i=1}^{n}V_{n,i}}\sum_{i=1}^{n}%
V_{n,i}\delta_{Z_{i}},
\]
with $\delta_{Z_{i}}$ the Dirac measure at $\{Z_{i}\}$. We also define a
weighted empirical error $\mathbb{P}_{n,V_{n}}\psi$ where $\mathbb{P}%
_{n,V_{n}}\psi$ stands for the usual notation of the expectation of $\psi$
with respect to $\mathbb{P}_{n,V_{n}}$. For $\mathbb{P}_{n,1_{n}}$, with
$1_{n}$ the binary vector of size $n$ with $1$ at every coordinate, we will
use the traditional notation $\mathbb{P}_{n}$. For a predictor trained on a
subsample, we define%
\[
\psi_{V_{n}}(.):=\Psi(.,\mathcal{D}_{V_{n}}).
\]

%
%
%

\noindent With the previous notations, notice that the predictor trained on
the learning set $\psi(.,\mathcal{D}_{n})\ $can be denoted by $\psi_{1_{n}%
}(.)$. We will allow the simpler notation $\psi_{n}(.)$. The learning set is
divided into two disjoint sets: the training set of size $n(1-p_{n})$ and the
test set of size $np_{n}$, where $p_{n}$ is the percentage of elements in the
test set. To represent the training set, we define $V_{n}^{tr}$ a random
binary vector of size $n$ independent of $\mathcal{D}_{n}$. $V_{n}^{tr}$ is
called the training vector. We define the test vector by $V_{n}^{ts}%
:=1_{n}-V_{n}^{tr}$ to represent the test set.

\noindent The distribution of $V_{n}^{tr}$ characterizes all the
cross-validation procedures described in the previous section (see e.g.
chapter 1). Using our notations, we can now define the cross-validation estimator.

\begin{definition}
[Cross-validation estimator]With the previous notations, the generalized
cross-validation error of $\psi_{n}$ denoted $\widehat{R}_{CV}(\psi_{n})$ is
defined by%
\[
\widehat{R}_{CV}(\psi_{n}):=\mathbb{E}_{V_{n}^{tr}}\mathbb{P}_{n,V_{n}^{ts}%
}(\psi_{V_{n}^{tr}}).
\]
\end{definition}

\noindent We will give here an example of distributions of
$V_{n}^{tr}$ to illustrate we retrieve cross-validation
procedures described previously. Leave-$\upsilon$-out
cross-validation is an elaborate and expensive version of
cross-validation. This procedure divides the data into two sets:
the training set of size $n-\upsilon$ and the test set of size
$\upsilon$. It then produces a predictor by training on the
training set and testing on the remaining test set. This is
repeated for all possible subsamples of $\upsilon$ cases, and the
observed errors are averaged to form the leave-$\upsilon$-out
estimate. Denote by $(\xi_{n,i}^{\upsilon})_{1\leq i\leq
\binom{{n}}{{\upsilon}}}$ the family of binary vectors of size
$n$ such that $\sum_{i=1}^{n}\xi_{n,i}^{\upsilon}=n-\upsilon$.

\begin{example}
[Leave-$\upsilon$-out cross-validation]%
\begin{align*}%
\Pr(V_{n}^{tr} &  =\xi_{n,1}^{\upsilon})=\frac{1}%
{\binom{{n}}{{\upsilon}}}%
\\
\Pr(V_{n}^{tr}
&  =\xi_{n,2}%
^{\upsilon})=\frac{1}%
{\binom{{n}}{{\upsilon}}}\\
&  \ldots\\
\Pr(V_{n}%
^{tr} &=\xi_{n,\binom{{n}}{{\upsilon}}%
}%
^{k}%
)=\frac{1}{\binom{{n}}{{\upsilon}}}.
\end{align*}%
\end{example}

\noindent For other examples, see chapter one.

\subsection{Definitions and notations of stability}

\noindent The basic idea is that an algorithm is stable at a
training set $\mathcal{D}_{n}$ if changing one point in
$\mathcal{D}_{n}$ yields only a small change in the output
hypothesis. Formally, a learning algorithm maps a weighted
training set into a predictor space. Thus, stability can be
translated into a Lipschitz condition for this mapping with high
probability.

\noindent To be more formal, following \cite{KUNIY02}, we define
a distance between two weighted empirical errors.

Let $\mathbb{P}_{n,V_{n}}$ and $\mathbb{P}_{n,U_{n}}$ be two
empirical measures on $\mathcal{Z}$ with respect to the binary
vectors $V_{n}$ and $U_{n}$. We do not assume their support to be
equal. The distance between
them is defined as their total variation, i.e. the number of points they do not have in common%
\[
||\mathbb{P}_{n,U_{n}}-\mathbb{P}_{n,V_{n}}||=\sup_{A\in\mathcal{P}%
(\mathcal{Z})}|(\mathbb{P}_{n,U_{n}}-\mathbb{P}_{n,V_{n}})(A)|.
\]

\begin{example}
In the case of leave-one-out (i.e. $\sum_{i=1}^{n}U_{n,i}=n-1$), we have%
\[
||\mathbb{P}_{n,U_{n}}-\mathbb{P}_{n}||=\frac{2}{n}.
\]

\noindent In the case of leave-$\nu$-out, we get%
\[
||\mathbb{P}_{n,U_{n}}-\mathbb{P}_{n}||=\frac{2\nu}{n}.
\]

\noindent In the general setting, it follows that%
\[
||\mathbb{P}_{n,U_{n}}-\mathbb{P}_{n}||=2p_{n}.%
\]

\end{example}

\noindent At least, we need a distance $d$ on the set $\mathcal{F}$. Let us
quote three important examples. Let $\psi_{1},\psi_{2}$ $\in\mathcal{F}$. The
uniform distance is defined by: $d_{\infty}(\psi_{1},\psi_{2})=\sup
_{Z\in\mathcal{Z}}|\psi_{1}(Z)-\psi_{2}(Z)|$, the $L_{1}$-distance by:
$d_{1}(\psi_{1},\psi_{2})=\mathbb{P}|\psi_{1}-\psi_{2}|$ , the error-distance
$d_{e}(\psi_{1},\psi_{2})=|\mathbb{P(}\psi_{1}-\psi_{2})|$. It is important to
notice that what matters here is not an absolute distance between the original
class of predictors $\mathcal{G}$ seen as functions but the distance with the
respect to the loss or/and the distribution $\mathbb{P}$. In particular, for
the $L_{1}$-distance, we do not care about the behavior of the original
predictors $\varphi_{1}$ and $\varphi_{2}$ outside the support of $\mathbb{P}%
$. At last, notice that we always have $d_{e}\leq d_{1}\leq d_{\infty}$.

\bigskip

\noindent We are now in position to define the different notions
of stability of a learning algorithm which cover notions
introduced by \cite{KUNIY02}. We begin with the notion of weak
stability. In essence, it says that for any given resampling
vectors, the distance between two predictors is controlled with
high probability by the distance between the resampling vectors.

\begin{definition}
[Weak stability]Let $\alpha,\lambda,(\delta_{n,p_{n}%
})_{n,p_{n}}$ be nonnegative real numbers. A learning algorithm
$\Psi$ is said to be weak $(\lambda,(\delta_{n,p_{n}})_{n,p_n},d)$
stable if for any training vector
$U_{n}$ whose sum is equal to $n(1-p_{n})$:%
\[
\Pr(d(\psi_{U_{n}},\psi_{n})\geq\lambda||\mathbb{P}_{n,U_{n}}-\mathbb{P}%
_{n}||^{\alpha})\leq\delta_{n,p_{n}}.
\]
\end{definition}

\noindent Notice that in the former definition $\Pr$ stands for $\mathbb{P}%
^{\otimes n}$. Indeed, $\psi_{n}$ is trained with $n$ observations, drawn
independently from $\mathbb{P}$. A stronger notion is to consider $\psi_{n}$
trained with $n-1$ observations drawn independently from $\mathbb{P}$ and an
additionnal general observation $z$. We consider the stronger notion of strong
stability. As a motivation, notice that algorithms such as Empirical Risk
Minimization with finite VC\ dimension (\cite{KUNIY02}) satisfies this property.

\begin{definition}
[Strong stability]Let $z\in\mathcal{Z}$. Let $\mathcal{D}_{n}=\mathcal{D}%
_{n-1}\cup\{z\}$ be a learning set. Let
$\lambda,(\delta_{n,p_{n}})_{n,p_{n}}$ be nonnegative real
numbers. A learning algorithm $\Psi$ is said to be strong
$(\lambda,(\delta_{n,p_{n}})_{n,p_n},d)$ stable if for any
training vector $U_{n}$ whose
sum is equal to $n(1-p_{n})$%
\[
\Pr(d(\psi_{U_{n}},\psi_{n})\geq\lambda||\mathbb{P}_{n,U_{n}}-\mathbb{P}%
_{n}||^{\alpha})\leq\delta_{n,p_{n}}.
\]
\end{definition}

\noindent What we have in mind for classical algorithms is $\delta_{n,p_{n}%
}=O_{n}(p_{n}\exp(-n(1-p_{n}))$. We can state the last definition in other
words. Let $V_{n}^{tr}$ be a training vector with distribution $\mathbb{Q}$
such that the number of elements in the training set is constant and equal to
$n(1-p_{n})$. Notice then that the former definition also implies that
$\sup_{U_{n}\in\text{support}(\mathbb{Q)}}\mathbb{P}(\frac{d(\psi_{U_{n}}%
,\psi_{n})}{||\mathbb{P}_{n,U_{n}}-\mathbb{P}_{n}||^{\alpha}}\geq\lambda
)\leq\delta_{n,p_{n}}$, where support$(\mathbb{Q)}$ stands for
the support of $\mathbb{Q}$. The previous notion stands for any
$U_{n}$ having the same support of $\mathbb{Q}$. A stronger
hypothesis would be that the previous probability stands
uniformly over $U_{n}$ in support$(\mathbb{Q)}$. This leads
formally to the notion of cross-validation stability. \noindent
As a motivation, notice that algorithms such as Lasso
(\cite{BTW07}) satisfies this property. To be more accurate, we
define

\begin{definition}
[Cross-validation weak stability]Let
$\mathcal{D}_{n}=(Z_{i})_{1\leq i\leq n}$ a learning set. Let
$V_{n}^{tr}$ a training vector with distribution $\mathbb{Q}$.
Let $\lambda,(\delta_{n,p_{n}})_{n,p_{n}}$ be nonnegative real
numbers. A learning algorithm $\Psi$ is said to be weak
$(\lambda,(\delta _{n,p_{n}})_{n,p_{n}},d,\mathbb{Q})$ stable if
it is weak $(\lambda
,(\delta_{n,p_{n}})_{n,p_{n}},d)$ stable and if:%
\[
\Pr(\sup_{U_{n}\in\text{support}(\mathbb{Q)}}\frac{d(\psi_{U_{n}},\psi_{n}%
)}{||\mathbb{P}_{n,U_{n}}-\mathbb{P}_{n}||^{\alpha}}\geq\lambda)\leq
\delta_{n,p_{n}}.
\]
\end{definition}

\noindent As before, we also define the following stronger notion

\begin{definition}
[Cross-validation strong stability]Let $z\in\mathcal{Z}$. Let $\mathcal{D}%
_{n}=\mathcal{D}_{n-1}\cup\{z\}$ a learning set. Let $V_{n}^{tr}$
be a cross-validation vector with distribution $\mathbb{Q}$. A
learning algorithm
$\Psi$ is said to be strongly $(\lambda,(\delta_{n,p_{n}})_{n,p_{n}%
},d,\mathbb{Q})$ stable if it is strong $(\lambda,(\delta_{n,p_{n}})_{n,p_{n}%
},d)$ stable and if:%
\[
\Pr(\sup_{U_{n}\in\text{support}(\mathbb{Q)}}\frac{d(\psi_{U_{n}},\psi_{n}%
)}{||\mathbb{P}_{n,U_{n}}-\mathbb{P}_{n}||^{\alpha}}\geq\lambda)\leq
\delta_{n,p_{n}}.
\]
\end{definition}

\begin{remark}
\label{Ch2:Rem1}If the cardinal of the support of $\mathbb{Q}$ is
denoted
$\kappa(n)$, then a learning algorithm which is weak $(\lambda,(\delta_{n,p_{n}%
})_{n,p_{n}},d,\mathbb{Q})$-stable is also strong $(\lambda,(\kappa
(n)\delta_{n,p_{n}})_{n},d,\mathbb{Q})$-stable.
\end{remark}

\noindent At last, we consider the special important case when $\delta
_{n,p_{n}}=0$. This is the case in particular for regularization networks
(\cite{BE01}).

\begin{definition}
[Sure stability]Notice that when $\delta_{n,p_{n}}=0$, the two notions
coincides and are called \textbf{sure stability}.
\end{definition}

\bigskip

\noindent As an example of strong stability, we develop the description of
\cite{FRE95} who introduced the algorithm.

\begin{example}
[Adaboost]We give an initial distribution $p^{1}$ and let
$w^{1}=p^{(1)}$ and $Z_{1}=1$. Let $\Phi$ be a learning
algorithm. Let $T$ the number of rounds.

For each $t=1...T:$

\begin{enumerate}
\item  Train the learning algorithm $\Phi$ on the learning set with distribution
$p^{(t)}$. The predictor obtained is denoted by $\varphi^{(t)}$.

\item  For each $i$, let $a_{i}^{t}=|\varphi^{t}(x_{i})-y_{i}|$, the error of
$\varphi^{(t)}$ on instance $i$.

\item  Let $\varepsilon_{t}=\sum_{i=1}^{m}p^{t}a_{i}^{t}$, the error rate of
$\varphi^{(t)}$ with respect to $p^{^{\prime}t)}$.

\item  Let $\beta_{t}=\frac{\varepsilon_{t}}{1-\varepsilon_{t}}$ and let
$\alpha_{t}=\ln(1/\beta_{t})$

\item  reweight the data: for all $i$, let $w_{i}^{(t+1)}=w_{i}^{(t)}\beta
_{t}^{1-a_{i}^{t}}$.

\item  Normalize the distribution: let $Z_{t+1}=\sum_{i=1}^{m}w_{i}^{(t+1)}$
and $p_{i}^{(t+1)}=w_{i}^{^{\prime}t+1)}/Z_{t+1}$
\end{enumerate}

\noindent The final output is
$H_{T}(x)=\sum_{s=1}^{T}\alpha_{s}\varphi^{(s)}(x)$.

\noindent \cite{KUNIY01} shows that under certain hypotheses,
Adaboost is strongly stable: suppose the  learner $\Phi$\
-$(\lambda,0,d_{\infty})$ stable  and other regularity
assumptions, then Adaboost with $T$ rounds is strong $(\lambda^{\ast}%
,(\delta_{n,p_{n}}^{\ast})_{n,p_{n}},d_{\infty})$ stable for some
$\lambda^{\ast}$ and $\delta_{n,p_{n}}^{\ast}$.
\end{example}

\noindent We give now an example that is surely stable introduced in \cite{BE01}.

\begin{example}
[Regularization networks ]Regularization networks are attractive for their
links with Support Vector Machines and their Bayesian interpretation. This
learning algorithm consists in finding a function $\varphi:\mathbb{R}%
^{d}\mathbb{\rightarrow R}$ in a space $H$ which minimizes the following
functional:%
\[
A(\varphi)=\frac{1}{n}\sum_{i=1}^{n}(Y_{i}-\varphi(X_{i}))^{2}+\lambda
||\varphi||_{H}^{2},
\]

\noindent with $||\varphi||_{H}$ the $L_{2}$ norm in the space $H$. $H$ is
chosen to be a reproducing kernel Hilbert Space (rkhs) with kernel $k$. $k$ is
supposed to be a symmetric function $k:\mathbb{R}^{d}\times\mathbb{R}^{d}%
\rightarrow\mathbb{R}$. \noindent In particular, we have the following
property (for a detailed introduction of rkhs, see \cite{ATE92})
\[
|f(x)|\leq||f||_{H}||k||_{H}.
\]
\noindent We slightly adapt the proof in \cite{BE01} to show that a
regularization network is surely stable:

\begin{theorem}
If $\Psi$ is a regularization network such that
$||k||_{H}\leq\kappa$ and $(y-\varphi(x))^{2}\leq
M\label{conditionBE}$, then $\Psi$ is $\frac
{4M\kappa^{2}}{n\lambda}$- surely stable with respect to the
distance $d_{\infty}$.
\end{theorem}

\noindent{\bfseries Proof}

\noindent Define $A^{i}(\varphi):=\frac{1}{n-1}\sum_{j\neq i}(Y_{j}%
-\varphi(X_{j}))^{2}+\lambda||\varphi||_{H}^{2}$ and $\mathcal{D}_{n}%
^{i}:=\mathcal{D}_{n}\backslash\{(X_{i},Y_{i})\}$. $\varphi_{\mathcal{D}%
_{n}^{i}}$ is the minimizer of $A^{i}$ over $H$ whereas $\varphi
_{\mathcal{D}_{n}}$ is the minimizer of $A$. Denote $g:=\varphi_{\mathcal{D}%
_{n}^{i}}-\varphi_{\mathcal{D}_{n}}$.\bigskip

\noindent For $t\in\lbrack0,1]$, we have
$A(\varphi_{\mathcal{D}_{n}})-A(\varphi_{\mathcal{D}_{n}}+tg)$ is
equal to  {\small
\[
\frac
{-2t}{n(n-1)}\sum_{j\neq i}(n-1)(\varphi_{\mathcal{D}_{n}}(x_{j}%
)-y_{j})g(x_{j})-\frac{2t}{n(n-1)}(n-1)(\varphi_{\mathcal{D}_{n}}(x_{i}%
)-y_{i})g(x_{i})-2t\lambda<\varphi_{\mathcal{D}_{n}},g>_{H}+t^{2}B(g)
\]
} \noindent with $B(g)$ the factor of $t^{2}$.\bigskip

\noindent In the same way, we get that $A^{i}(\varphi_{\mathcal{D}_{n}^{i}})-A^{i}(\varphi_{\mathcal{D}_{n}^{i}%
}-tg)$ is equal to{\small
\[
\frac{2t}{n(n-1)}\sum_{j\neq i}(n-1)(\varphi_{\mathcal{D}_{n}^{i}}%
(x_{j})-y_{j})g(x_{j})+\frac{2t}{n(n-1)}\sum_{j\neq i}(\varphi_{\mathcal{D}%
_{n}^{i}}(x_{j})-y_{j})g(x_{j})+2t\lambda<\varphi_{\mathcal{D}_{n}^{i}}%
,g>_{H}+t^{2}B^{n}(g).
\]
}

\bigskip\

\noindent By definition of $A$ and $A^{i}$, we have $A(\varphi
_{\mathcal{D}_{n}})-A(\varphi_{\mathcal{D}_{n}}+tg)\leq0$ and $A^{i}%
(\varphi_{\mathcal{D}_{n}^{i}})-A^{i}(\varphi_{\mathcal{D}_{n}^{i}}+tg)\leq0$.
Thus, we get by summing these two inequalities, dividing by $\frac{2t}%
{n(n-1)}$ and making $t\rightarrow0$.%

\[
\sum_{j\neq i}({\small n-1)g}^{2}{\small (x}_{j}{\small )+}\sum_{j\neq
i}\left[  {\small (\varphi_{\mathcal{D}_{n}^{i}}(x}_{j}{\small )-y}%
_{j}{\small )g(x}_{j}{\small )-(\varphi}_{\mathcal{D}_{n}}{\small (x}%
_{i}{\small )-y}_{i}{\small )g(x}_{i}{\small )}\right]  +n(n-1)||g||_{H}%
^{2}\leq0
\]

\noindent which leads to
\[
n(n-1)||g||_{H}^{2}\leq\sum_{j\neq i}\left[  {\small (\varphi}_{\mathcal{D}%
_{n}}{\small (x}_{i}{\small )-y}_{i}{\small )g(x}_{i}{\small )-(\varphi
_{\mathcal{D}_{n}^{i}}(x}_{j}{\small )-y}_{j}{\small )g(x}_{j}{\small )}%
\right]  \leq2(n-1)\sqrt{M}\kappa||g||_{H}%
\]

\noindent by assumptions. \bigskip

\noindent Thus, we have%
\[
||g||_{H}\leq2(n-1)\sqrt{M}\kappa||g||_{H}/n\lambda
\]

\noindent and also, for all $x,y$%

\[
|({\small \varphi}_{\mathcal{D}_{n}}{\small (x)-y})^{2}-({\small \varphi
}_{\mathcal{D}_{n}^{i}}{\small (x)-y})^{2}|\leq2\sqrt{M}|{\small \varphi
}_{\mathcal{D}_{n}}{\small (x)-\varphi}_{\mathcal{D}_{n}^{i}}{\small (x)}%
|\leq4M\kappa^{2}/n\lambda.
\]

$\Box$
\end{example}

\noindent Another popular example is given by the k-nearest neighbors which
are strongly stably with respect to $d_{1}$.

\begin{example}
[k-nearest neighbors ]In the $k$-nearest rule, the machine learning is a
function of $X$ and of the $k$ nearest observations to $X$ from $(X_{1}%
,...,X_{n})$ and of the corresponding $(Y_{1},...,Y_{n})$. Because
there may be ties in determining the $k$ nearest neighbors, we
use an independent sequence $(Z,Z_{1},...,Z_{n})$ of i.i.d uniform
random variables in $[0,1]$. $X_{j}$ is nearer $X_{i}$ to $X$ if:

\begin{enumerate}
\item $||X_{j}-X||<||X_{j}-X||$ or

\item $||X_{j}-X||=||X_{j}-X||$ and $|Z_{j}-Z|<|Z_{i}-Z|$, or

\item $||X_{j}-X||=||X_{j}-X||$ and $Z_{j}=Z_{i}$ and $j<i$.
\end{enumerate}

\noindent The last event does not count since its has zero probability.

\noindent Denote $\gamma_{d}$ the maximum number of distinct points in
$\mathbb{R}^{d}$ that share the same nearest neighbor. It can be shown that
$\gamma_{d}\leq3^{d}-1$ and other lower and upper bounds can be found in
\cite{ROG63}. Recall the following lemma from \cite{DEWA79}: suppose
$(X_{1},Z_{1}),\ldots,(X_{n},Z_{n})$ is the sequence obtained from the data by
omitting the $Y_{1},...,Y_{n}$. If, for each $j$, the nearest neighbor to
$(X_{j},Z_{j})$ is found from $(X_{1},Z_{1}),\ldots,(X_{j-1},Z_{j-1}%
),(X_{j+1},Z_{j+1}),...,(X_{n},Z_{n})$. Then no point
$(X_{i},Y_{i})$ can be the nearest neighbors to more than
$\gamma_{d}+2$ of the remaining points.

\noindent We can derive the next result following the proofs in \cite{DEWA79}.

\begin{theorem}
Let $\mathcal{D}_{n}:=((X_{1},Z_{1},Y_{1}),\ldots,(x,z,y))$ be a learning set.
Suppose $\Phi$ is a $k$ local rule. Then we have for all $\varepsilon>0$,%
\[
\Pr(E_{X,Y,Z}|L(Y,\Phi((X,Z),\mathcal{D}_{n}))-L(Y,\Phi((X,Z),\mathcal{D}%
_{n}^{i}))|\geq\varepsilon)\leq6\exp(\frac{-(n-1)\varepsilon^{3}}%
{54k(\gamma_{d}+2)})
\]

\noindent with
$\mathcal{D}_{n}^{i}:=\mathcal{D}_{n}\backslash\{(X_{i},Y_{i},Z_{i})\}$
and $i$ a fixed index.
\end{theorem}

\bigskip

\noindent It says that the $k$ nearest rule satisfies strong stability
property with respect $d_{1}$ and $||\mathbb{P}_{n,U_{n}}-\mathbb{P}%
_{n}||^{\alpha}$ with $\alpha<1/3.$

\noindent{\bfseries Proof}\bigskip

\noindent Consider one local rule first.

\bigskip

\noindent Let $m$ be an integer. Consider an independent identically
distributed ghost sample
\[
((X_{n+1},Y_{n+1},Z_{n+1}),\ldots,(X_{n+m},Y_{n+m},Z_{n+m})).
\]
\bigskip\noindent Denote $\mathcal{T}_{n+m}:=((X_{1},Y_{1},Z_{1}%
),\ldots,(X_{n+m},Y_{n+m},Z_{n+m}))$ and $\mathcal{T}_{n+m}^{j}:=T_{n+m}%
\backslash\{(X_{j},Y_{j},Z_{j})\}.$

\begin{itemize}
\item $L_{1}:={\small E}_{X,Y,Z}{\small |L(Y,\Phi((X,Z),}\mathcal{D}%
_{n}{\small ))-L(Y,\Phi((X,Z),}\mathcal{D}_{n}^{i}{\small ))|}$

\item $L_{2}:=\frac{1}{m}\sum_{j=1}^{m}{\small |L(Y}_{n+j}{\small ,\Phi
((X}_{n+j}{\small ,Z}_{n+j}{\small ),}\mathcal{D}_{n}{\small ))-L(Y}%
_{n+j}{\small ,\Phi((X}_{n+j}{\small ,Z}_{n+j}{\small ),\mathcal{D}_{n}%
^{i}))|}$

\item $L_{3}:=\frac{1}{m}\sum_{j=1}^{m}{\small |L(Y}_{n+j}{\small ,\Phi
((X}_{n+j}{\small ,Z}_{n+j}{\small ),}\mathcal{D}_{n}{\small ))-L(Y}%
_{n+j}{\small ,\Phi((X}_{n+j}{\small ,Z}_{n+j}{\small ),}\mathcal{T}%
_{n+m}^{n+j}{\small ))|}$

\item $L_{4}:=\frac{1}{m}\sum_{j=1}^{m}{\small |L(Y}_{j}{\small ,\Phi
((X}_{n+j}{\small ,Z}_{n+j}{\small ),}\mathcal{D}_{n}^{i}{\small ))-L(Y}%
_{n+j}{\small ,\Phi((X}_{n+j}{\small ,Z}_{n+j}{\small ),}\mathcal{T}%
_{n+m}^{n+j}{\small ))|}$.\bigskip
\end{itemize}

\noindent We have
\[
\Pr(L_{1}\geq3\varepsilon)\leq\Pr(L_{1}-L_{2}\geq\varepsilon)+\Pr(L_{3}%
\geq\varepsilon)+\Pr(L_{4}\geq\varepsilon).
\]
\bigskip$_{\noindent}$By Hoeffding's inequality we have $\Pr(L_{1}-L_{2}%
\geq\varepsilon)\leq\exp(-2m\varepsilon^{2})$.\bigskip

\noindent Now we get for the second term%
\begin{align*}
\Pr(L_{3}    \geq\varepsilon)&\leq\Pr(\frac{1}{m}\sum_{j=1}^{m}1_{{\small \Phi
((X}_{n+j}{\small ,Z}_{n+j}{\small ),}\mathcal{D}_{n}{\small )\neq\Phi
((X}_{n+j}{\small ,Z}_{n+j}{\small ),}\mathcal{T}_{n+m}^{n+j}{\small )}}%
\geq\varepsilon)\\
&  \leq\Pr(\frac{1}{m}\sum_{j=1}^{m}1_{{\small A(n+j)}}\geq\varepsilon),
\end{align*}

\noindent with ${A(n+j)}$ the event that the nearest neighbor of
${\small (X}_{n+j}{\small ,Z}_{n+j}{\small )}$ from
$\mathcal{T}_{n+m}^{n+j}$ is attained in the ghost sample
$\mathcal{T}_{n+m}^{n+j}\backslash \mathcal{D}_{n}$.

\bigskip

$_{\noindent}$From \cite{DEWA79}, we have, if $(\gamma_{d}+2)m<(n+m)\varepsilon/2$,
\[
\Pr(\frac{1}{m}\sum_{j=1}^{m}1_{{\small A(n+j)}}\geq\varepsilon)\leq
2\exp(-2m(\varepsilon/2)^{2})
\]
\noindent  In the same way, we find
that $\Pr(L_{4}\geq\varepsilon)\leq2\exp(-2m(\varepsilon/2)^{2})$ if
$(\gamma_{d}+2)m<(n-1+m)\varepsilon/2$

\bigskip

\noindent Taking $m=\frac{(n-1)\varepsilon}{\gamma_{d}+2}$, we obtain
\[
\Pr(L_{3}\geq\varepsilon)\leq2\exp(\frac{-(n-1)\varepsilon^{3}}{2(\gamma
_{d}+2)})
\]
\noindent and $\Pr(L_{3}\geq\varepsilon)\leq2\exp(\frac{-(n-1)\varepsilon^{3}%
}{2(\gamma_{d}+2)})$.

\bigskip

\noindent For an arbitray $k$, it is sufficient to replace $(\gamma_{d}+2)$ by
$k(\gamma_{d}+2)$.

$\Box$
\end{example}

\bigskip

\noindent A last popular example is given by the Lasso which is strongly
stable with respect to $d_{1}$.

\begin{example}
[Lasso]We follow \cite{BTW07} who defines Lasso-type methods in
the following way. Let $((X_{1},Y_{1}),...,(X_{n},Y_{n}))$ be a
sample of i.i.d. pairs distributed as
$(X,Y)\in(\mathcal{X}\mathcal{,}\mathbb{R})$, where $\mathcal{X}$
is a borel subset of $\mathbb{R}^{d}$. We denote by $\mu$ the
distribution of $X$ on $\mathcal{X}$. Let $f(X)=\mathbb{E}(Y|X)$
be the \ unknown regression function and
$\mathcal{F}_{M}=\{f_{1},...,f_{M}\}$ be a dictionary of
real-valued functions $f_{j}$ that are defined on $\mathcal{X}$.
We use a data dependent $l_{1}$-penalty. Formally, for any
$\lambda =(\lambda_{1},...,\lambda_{M})\in \mathbb{R}^{M}$ ,
define $f_{\lambda}(x)=\sum _{j=1}^{M}\lambda_{j}f_{j}(x)$. Then
the penalized least squares estimator of
$\lambda$ is%
\[
\hat{\lambda}=\arg\min\{1/n\sum_{i=1}^{n}(Y_{i}-f_{\lambda}(X_{i}%
))^{2}+pen(\lambda)\}
\]

\noindent where
\[
pen(\lambda)=2\sum_{j=1}^{M}\omega_{n,j}|\lambda_{j}|\text{ with }\omega
_{n,j}=r_{n,M}||f_{j}||_{n}%
\]

\noindent where $||g||_{n}^{2}=n^{-1}\sum_{i=1}^{n}g^{2}(X_{i})$
for the squared empirical empirical $L_{2}$ norm of any function
$g:\mathcal{X\rightarrow}R$. The tuning sequence $r_{n,M}>0$ is
defined by $r_{n,M}:=A\sqrt{\log(M)/n}$ for $A$ large enough.
Then we have $\hat{f}_{n}=f_{\hat{\lambda}}$

\noindent Define%
\[
M(\lambda)=\sum_{j=1}^{M}I_{\{\lambda_{j}\neq0\}}%
\]
the number of non-zero coordinates of $\lambda$.

\noindent We recall the definition of weak sparsity in
\cite{BTW07}. Let $C_{f}>0$ be a constant depending only on $f$
and
\[
\Lambda=\{\lambda\in R^{M}:||f_{\lambda}-f||^{2}\leq C_{f}r_{n,M}^{2}%
M(\lambda)\}
\]

\noindent where
\[
||g||^{2}=\int_{\mathcal{X}}g^{2}(x)\mu(dx)
\]

\noindent If $\Lambda$ is not empty, $f$ has the weak sparsity
property relative to the dictionary $\{f_{1},...,f_{M}\}$.

\noindent We have then the following theorem

\begin{theorem}
Assume the general assumptions (A1)-(A3) and consider the notations in \cite{BTW07}.
Then, for all $\lambda\in\Lambda,$%
\[
\Pr(|E_{X,Y}(Y-\hat{f}_{n}(X))^{2}-E_{X,Y}(Y-\hat{f}_{n-1}(X))^{2}%
|>2B_{1}\kappa_{M}^{-1}r_{n,M}^{2}M(\lambda))\leq2\pi_{n-1,M}(\lambda)
\]

\noindent with $\pi_{n-1,M}(\lambda)$ a small probability defined
in \cite{BTW07}.

\noindent In other words, the Lasso-type algorithm is weakly
stable with respect to $d_{e}$ and $||.||_{1}$.
\end{theorem}

\noindent {\bfseries Proof}

\bigskip

According to theorem 2.1. in \cite{BTW07}, we have:
\[
\Pr(E_{X,Y}|\hat{f}_{n}(X))-f(X)|^{2}\leq B_{1}\kappa_{M}^{-1}r_{n,M}%
^{2}M(\lambda))\geq1-\pi_{n,M}(\lambda).
\]

Thus, denote $\pi:=\Pr(|E_{X,Y}(Y-\hat{f}_{n}(X))^{2}-E_{X,Y}(Y-\hat{f}%
_{n-1}(X))^{2}|>2B_{1}\kappa_{M}^{-1}r_{n,M}^{2}M(\lambda))$. We
obtain:{\tiny  }%
\begin{align*}
\pi&=\Pr(|E_{X}(f(X)-\hat{f}_{n}(X))^{2}-E_{X,Y}(f(X)-\hat{f}_{n-1}%
(X))^{2}|>2B_{1}\kappa_{M}^{-1}r_{n,M}^{2}M(\lambda))\\
&  \leq\Pr(E_{X}(f(X)-\hat{f}_{n}(X))^{2}>B_{1}\kappa_{M}^{-1}r_{n,M}%
^{2}M(\lambda)) \\
& +\Pr(E_{X}(f(X)-\hat{f}_{n-1}(X))^{2}>B_{1}\kappa_{M}%
^{-1}r_{n,M}^{2}M(\lambda))\\
&  \leq2\pi_{n-1,M}(\lambda).
\end{align*}

$\Box$
\end{example}

\bigskip

\noindent As seen in the following table, we retrieve with those notations the
different notions of stability introduced by \cite{DEWA79}, \cite{KEA95} and
also \cite{BE01}, \cite{KUNIY02}.%

\[%
\begin{tabular}
[c]{|l|l|l|l|}\hline {\tiny stability $\backslash$  distance} &
${\tiny d}_{\infty}$ & ${\tiny d}_{1}$ & ${\tiny d}_{e}$\\\hline
{\tiny Weak} &
\begin{tabular}
[c]{l}%
{\tiny weak }${\tiny (\lambda,\delta)}$ {\tiny hypothesis stability}\\
{\tiny \cite{KUNIY02}}%
\end{tabular}
&
\begin{tabular}
[c]{l}%
{\tiny weak }${\tiny (\lambda,\delta)}$ ${\tiny L}_{{\tiny 1}}$%
{\tiny stability}\\
{\tiny \cite{KUNIY02}}%
\end{tabular}
&
\begin{tabular}
[c]{l}%
{\tiny weak }${\tiny (\lambda,\delta)}$ {\tiny error stability}\\
{\tiny \cite{KUNIY02}}%
\end{tabular}
\\\hline
{\tiny Strong} &
\begin{tabular}
[c]{l}%
{\tiny strong }${\tiny (\lambda,\delta)}$ {\tiny hypothesis stability}\\
{\tiny \cite{KUNIY02}\cite{DEWA79}}%
\end{tabular}
&
\begin{tabular}
[c]{l}%
{\tiny strong }${\tiny (\lambda,\delta)}$ ${\tiny L}_{{\tiny 1}}%
${\tiny stability}\\
{\tiny \cite{KUNIY02}}%
\end{tabular}
&
\begin{tabular}
[c]{l}%
{\tiny strong }${\tiny (\lambda,\delta)}$ {\tiny error stability}\\
{\tiny \cite{KUNIY02}}%
\end{tabular}
\\\hline
{\tiny Sure Stability} &
\begin{tabular}
[c]{l}%
{\tiny uniform stability}\\
{\tiny \cite{BE01}}%
\end{tabular}
& {\tiny \cite{DEWA79}} &
\begin{tabular}
[c]{l}%
{\tiny error stability}\\
{\tiny \cite{KEA95}}%
\end{tabular}
\\\hline
\end{tabular}
\ \
\]

\noindent To motivate this approach, we also quote a list of class of
predictors satisfying the previous stability conditions.%

\[%
\begin{tabular}
[c]{|l|l|l|l|}\hline
{\small stability distance} & ${\small d}_{\infty}$ & ${\small d}_{1}$ &
${\small d}_{e}$\\\hline
{\small Weak} &  &  & {\small Lasso}%
\\\hline
{\small Strong} &  {\small Adaboost (}{\tiny \cite{KUNIY02})} &
\begin{tabular}
[c]{l}%
{\small -ERM (}{\tiny \cite{KUNIY02})}\\
-$k${\small -nearest rule}%
\end{tabular}
&
\begin{tabular}
[c]{l}%
{\small Bayesian algorithm}\\
{\small \cite{KEA95}}%
\end{tabular}
\\\hline
{\small Uniform} & {\small Regularization networks } &  &
\\\hline
\end{tabular}
\ \
\]

\begin{remark}
We omit other weaker definition of stability such as defined in \cite{BE01},
\cite{DEWA79}, and \cite{KUNIY02}. They consider bounds on the first moment of
$\mathbb{E}_{\mathcal{D}_{n}}d(\psi_{U_{n}},\psi_{n})$ instead of probability
bounds. Under these assumptions, they obtain polynomial upper bounds on
$\Pr\mathbb{(}|\widehat{R}_{CV}-\widetilde{R}_{n}|\geq\varepsilon)$. It is
would be interesting to explore the behaviour of cross-validation estimates
under these hypotheses. However, this cannot be done with the techniques
presented in this paper and is left to further investigation.
\end{remark}

\bigskip

\noindent The main notations and definitions are summarized in the next table:

\begin{table}[th]
\begin{center}%
\begin{tabular}
[c]{lll}\hline
Name & Notation & Definition\\\hline\hline
Risk or generalization error & $\widetilde{R}_{n}$ & $E_{P}[L(Y,\phi
(X,D_{n}))\mid D_{n}]$\\
Resubstitution error & $\widehat{R}_{n}$ & $\frac{1}{n}\sum_{i=1}^{n}%
L(Y_{i},\phi_{n}(X_{i},D_{n}))$\\
Cross-validation error & $\widehat{R}_{CV}$ & $E_{V_{n}^{tr}}P_{n,V_{n}^{ts}%
}\psi_{V_{n}^{tr}}$\\\hline
\end{tabular}
\end{center}
\caption{Main notations}%
\end{table}

\section{Results for risk assessment for stable algorithms}

\label{results}

\noindent Our goal is now to derive upper bounds for the probability that the
distance between the cross-validation estimator and the generalization error
is greater than $\varepsilon\geq0$: $\Pr\mathbb{(}|\widehat{R}_{CV}%
-\widetilde{R}_{n}|\geq\varepsilon)$.

\subsection{Hypotheses $\mathcal{H}$}

\noindent Let $\mathcal{D}_{n}$ be a learning set of size $n$. Let $V_{n}%
^{tr}\sim$ $\mathbb{Q}$ be a training vector independent of $\mathcal{D}_{n}$
such that the cross-validation is symmetric -i.e. $\Pr(V_{n,i}^{tr}=1)$ is a
constant independent of $i$ --and the number of elements in the training set
is equal to $np_{n}$. Let $d$ be a distance among $d_{e}%
,d_{1},d_{\infty}$. At last, we suppose that the loss function $L$ is bounded
by $1$. We derive the following general results that stands for general
cross-validation procedures and stable algorithms.

\subsection{Strong stability}

\noindent We state two results according to the class of stability. We will
use the definition of strong difference bounded introduced by \cite{KUT02} and
a corollary of his main theorem inspired by \cite{McD89}.

\begin{definition}
[Kutin\cite{KUT02}]Let $\Omega_{1},\ldots,\Omega_{n}$ be probability spaces.
Let $\Omega=\prod_{k=1}^{n}\Omega_{k}$ and let $X$ a random variable on
$\Omega$. We say that $X$ is strongly difference bounded by $(b,c,\delta)$ if
the following holds: there is a \textquotedblright bad\textquotedblright
\ subset $B\subset\Omega$, where $\delta=\mathbb{P}(B)$. If $\omega
,\omega^{\prime}\in\Omega$ differ only in $k$-th coordinate, and $\omega\notin
B$, then%

\[
|X(\omega)-X(\omega^{\prime})| \leq c.
\]

Furthermore, for any $\omega, \omega^{\prime}\in\Omega$,
\[
|X(\omega)-X(\omega^{\prime})| \leq b.
\]
\end{definition}

\noindent We will need the following theorem. It says in substance that a
strongly difference bounded function of independent variables is closed to its
expectation with high probability.

\begin{theorem}
[Kutin\cite{KUT02}]\label{ch2:Kutin strong} Let
$\Omega_{1},\ldots,\Omega_{n}$ be probability spaces. Let
$\Omega=\prod_{k=1}^{n}\Omega_{k}$ and let $X$ a random variable
on $\Omega$, which is strongly difference bounded by
$(b,c,\delta)$. Assume $b\geq c\geq0$ and $\alpha'>0$. Let
$\mu=\mathbb{E}(X)$.
Then, for any $\tau>0,\alpha'>0$,%

\[
\Pr(X-\mu\geq\tau)\leq2(\exp(-\frac{\tau^{2}}{8n(c+b\alpha^{^{\prime}})^{2}%
})+\frac{n}{\alpha^{^{\prime}}}\delta).
\]
\end{theorem}

\noindent We are now in position to derive

\begin{theorem}
[Cross-validation  strong stability]\label{th:CV strong}Suppose that
$\mathcal{H}$ holds. Let $\Psi$ a machine learning which is strong
$(\lambda,(\delta_{n,p_{n}})_{n,p_{n}},d,\mathbb{Q})$ stable. Then, for all
$\varepsilon\geq0$,%

\[
\Pr\mathbb{(}|\widehat{R}_{CV}-\widetilde{R}_{n}|\geq\varepsilon
+\lambda(2p_{n})^{\alpha})\leq2\exp(-2np_{n}\varepsilon^{2})+\delta_{n,p_{n}%
}.
\]

\noindent Furthermore, if $d$ is the uniform distance $d_{\infty}$, then we
have for all $\varepsilon\geq0$:%

\[
\Pr\mathbb{(}|\widehat{R}_{CV}-\widetilde{R}_{n}|\geq\varepsilon
+\delta_{n(1-p_{n})}+\lambda(2p_{n})^{\alpha})\leq4(\exp(-\frac{\varepsilon
^{2}}{8n(5\lambda(2p_{n})^{\alpha}+\alpha^{^{\prime}})^{2}})+\frac{n}%
{\alpha^{^{\prime}}}\delta_{n,p_{n}}^{^{\prime}}),
\]

\noindent with $\delta_{n,p_{n}}^{^{\prime}}=\delta_{n,p_{n}}+(n+1)\delta
_{n+1,1/(n+1)}.$

\noindent Thus, if we choose $\alpha=5 \lambda
(2np_{n})^{\alpha}$,
\[
\Pr\mathbb{(}|\widehat{R}_{CV}-\widetilde{R}_{n}|\geq\varepsilon
+\delta_{n,p_{n}}+\lambda(2p_{n})^{\alpha})\leq4(\exp(-\frac{\varepsilon
^{2}}{8(10\lambda)^{2}n(2p_{n})^{2\alpha}})+\frac{n}{5\lambda(2p_{n})^{\alpha
}}\delta_{n,p_{n}}^{^{\prime}}).
\]
\end{theorem}

\noindent{\bfseries Proof}

\begin{enumerate}
\item \noindent For the general case, denote $B$ the bad subset, i.e.
$B=\{\sup_{U_{n}\in\text{support}(\mathbb{Q)}}\frac{d(\psi_{U_{n}},\psi_{_{n}%
})}{||\mathbb{P}_{n,U_{n}}-\mathbb{P}_{n}||_{\alpha}}\geq\lambda\}$. Since
$\Psi$ is strong $(\lambda,(\delta_{n,p_{n}})_{n,p_{n}},d,\mathbb{Q})$ stable,
we have $\Pr\mathbb{(}B)\leq\delta_{n,p_{n}}$. It is sufficient to split
$|\widehat{R}_{CV}-\widetilde{R}_{n}|$ according to a benchmark, namely
$\overline{R}_{n(1-p_{n})}:=\mathbb{E}_{V_{n}^{tr}}\mathbb{P}\psi_{V_{n}^{tr}%
}$. Thus, we get
\[
\Pr\mathbb(|\widehat{R}_{CV}-\widetilde{R}_{n}|\geq\varepsilon+\lambda
(2p_{n})^{\alpha})\leq\Pr\mathbb{(}|\widehat{R}_{CV}-\overline{R}_{n(1-p_{n})}%
|\geq\varepsilon)+\Pr\mathbb{(}|\overline{R}_{n(1-p_{n})}-\widetilde{R}%
_{n}|\geq\lambda(2p_{n})^{\alpha})
\]

\noindent The first term can be bounded by conditional Hoeffding inequality
(see chapter 1). Thus, we obtain
\[
\Pr\mathbb{(}|\widehat{R}_{CV}-\overline{R}_{n(1-p_{n})}|\geq\varepsilon
)\leq2\exp(-2np_{n}\varepsilon^{2}).
\]

\bigskip

\noindent For the second term, notice that:
\[
|\overline{R}_{n(1-p_{n})}-\widetilde{R}_{n}|=|\mathbb{E}_{V_{n}^{tr}%
}\mathbb{P}\psi_{V_{n}^{tr}}-\mathbb{P}\psi_{n}|\leq\mathbb{E}_{V_{n}^{tr}%
}|\mathbb{P}\psi_{V_{n}^{tr}}-\mathbb{P}\psi_{n}|.
\]
\noindent Recall that for any $d\in\{d_{e},d_{1},d_{\infty}\}$, we have
$|\mathbb{P}\psi_{V_{n}^{tr}}-\mathbb{P}\psi_{n}|\leq d(\psi_{V_{n}^{tr}}%
,\psi_{n})$ and $||\mathbb{P}_{n,V_{n}^{tr}}-\mathbb{P}_{n}||_{\alpha}%
=(2p_{n})^{\alpha}$.

\bigskip

\noindent Thus, since $\Psi$ is strong $(\lambda,(\delta_{n})_{n})$ stable, we
have
\begin{align*}
\Pr\mathbb(|\overline{R}_{n(1-p_{n})}-\widetilde{R}_{n}|\geq\lambda
(2p_{n})^{\alpha}) &  \leq \Pr(\sup_{V_{n}^{tr}\in \text{support}(\mathbb{Q})}%
d(\psi_{V_{n}^{tr}},\psi_{n})/||\mathbb{P}_{n,V_{n}^{tr}}-\mathbb{P}%
_{n}||_{\alpha}\geq\lambda)\\
&=\Pr\mathbb{(}B)\leq\delta_{n,p_{n}}%
\end{align*}

\item  In the particular case, when $d=d_{\infty}$, the most stable notion of
stability, we can obtain a stronger result. For this, we recall two very
useful results.
\end{enumerate}

\noindent We proceed in three steps as in
\cite{BE02},\cite{KUNIY02} by using a bounded difference
inequality

\begin{itemize}
\item  first, we show that the expectation of $\widehat{R}_{CV}-\widetilde
{R}_{n}$ is small,

\item  secondly, we show that the function $\widehat{R}_{CV}-\widetilde{R}%
_{n}$ seen as a function $f$ of $Z_{1},Z_{2},\ldots,Z_{n}$ is strongly
difference bounded, i.e.: with high probability, there exists constants
$c_{1},\ldots,c_{n}$ such that we have for all $i$, for all $z\in
\mathcal{Z}$, $$|f(Z_{1},\ldots,Z_{i},\ldots,Z_{n})-f(Z_{1},\ldots,Z_{i-1}%
,z,Z_{i+1},\ldots,Z_{n})|\leq c_{i},$$

\item  use theorem \ref{ch2:Kutin strong} with the first two points,

\item  at least, use arguments of symmetry to conclude.
\end{itemize}

\begin{enumerate}
\item \textbf{The expectation of }$\widehat{R}_{CV}-\widetilde{R}_{n}$
\textbf{is small}

\bigskip

\noindent Let us denote $\mathbf{v}_{n}^{tr},\mathbf{v}_{n}^{ts}$ fixed
training and test vectors.%

\[
\mathbb{P}^{\otimes n}(\widehat{R}_{CV}-\widetilde{R}_{n})=\mathbb{P}^{\otimes
n}(\mathbb{E}_{V_{n}^{tr}}\mathbb{P}_{n,V_{n}^{ts}}\psi_{V_{n}^{tr}%
}-\mathbb{P}\psi_{n})=\mathbb{P}^{\otimes n}\mathbb{P(}\psi_{\mathbf{v}%
_{n}^{tr}}-\psi_{n})
\]
\noindent since%
\[
\mathbb{P}^{\otimes n}\mathbb{E}_{V_{n}^{tr}}\mathbb{P}_{n,V_{n}^{ts}}%
\psi_{V_{n}^{tr}}=\mathbb{E}_{V_{n}^{tr}}\mathbb{P}^{\otimes n}\mathbb{P}%
_{n,V_{n}^{ts}}\psi_{V_{n}^{tr}}=\mathbb{E}_{V_{n}^{tr}}\mathbb{P}^{\otimes
n}\mathbb{P}\psi_{V_{n}^{tr}}=\mathbb{P}^{\otimes n}\mathbb{P}\psi
_{\mathbf{v}_{n}^{tr}}%
\]
\noindent where the first equality comes from the linearity of expectation,
the second from the fact that $\mathbb{P}_{n,V_{n}^{tr}}$ are independent of
$\mathbb{P}_{n,V_{n}^{ts}}$, and the third from the i.i.d. nature of
$(Z_{i})_{i}$.

\bigskip

\noindent Recall that $\mathbb{P}(\psi_{\mathbf{v}_{n}^{tr}}-\psi_{n})\leq
d(\psi_{\mathbf{v}_{n}^{tr}},\psi_{n})$ where $d$ stands indifferently for
$d_{1},d_{e}$ or $d_{\infty}$. Thus, $\mathbb{P}^{\otimes n}\mathbb{P(}%
\psi_{\mathbf{v}_{n}^{tr}}-\psi_{n})\leq\mathbb{P}^{\otimes n}d(\psi
_{\mathbf{v}_{n}^{tr}},\psi_{n})$. By conditioning according to the small
values of $d(\psi_{\mathbf{v}_{n}^{tr}},\psi_{n})$,we obtain%

\begin{align*}
\mathbb{P}^{\otimes n}d(\psi_{\mathbf{v}_{n}^{tr}},\psi_{n})  &
=\mathbb{P}^{\otimes n}(d(\psi_{\mathbf{v}_{n}^{tr}},\psi
_{n})|B)\mathbb{P}^{\otimes n}(B)+\mathbb{P}^{\otimes n%
}(d(\psi_{\mathbf{v}_{n}^{tr}},\psi_{n})|B^{\subset})(1-\mathbb{P}%
^{\otimes n}(B))\\
&  \leq1\times\delta_{n,p_{n}}+\lambda\mathbb{P}^{\otimes n%
}||\mathbb{P}_{n,\mathbf{v}_{n}^{tr}}-\mathbb{P}_{n}||_{\alpha}\times
(1-\delta_{n,p_{n}})=\delta_{n,p_{n}}+\lambda(2p_{n})^{\alpha}(1-\delta
_{n,p_{n}})
\end{align*}

\noindent Eventually, we get $\mathbb{P}^{\otimes n}(\widehat{R}%
_{CV}-\widetilde{R}_{n})\leq\delta_{n,p_{n}}+\lambda(2p_{n})^{\alpha}$.

\item $\widehat{R}_{CV}-\widetilde{R}_{n}$ \textbf{is difference bounded with
high probability}

\bigskip

\noindent Denote
$f(Z_{1},Z_{2},\ldots,Z_{n}):=\widehat{R}_{CV}-\widetilde
{R}_{n}$. Let $z$ $\in$ $\mathcal{Z}$. Let $\mathcal{D}_{n+1}=
\mathcal{D}_{n+1} \cup \{ z\}$. Now denote $B=B_{1}\cup B_{2}$
where
$$B_{1}=\{\sup_{U_{n}\in\text{support}(\mathbb{Q)}}\frac{d(\psi_{U_{n}}%
,\psi_{n})}{||\mathbb{P}_{n,U_{n}}-\mathbb{P}_{n}||_{\alpha}}\geq\lambda\}$$

\noindent  and
$$B_{2}=\{\sup_{1\leq i\leq n+1}\frac{d(\psi_{e_{n+1}^{i}},\psi_{n+1}%
)}{||\mathbb{P}_{n+1,e_{n+1}^{i}}-\mathbb{P}_{n+1}||_{\alpha}}\geq\lambda\}$$

\noindent with $e_{n+1}^{i}$ the binary of size $n+1$ equal to
$0$ everywhere except on the $i$-th coordinate
$e_{n+1,k}^{i}:=1_{(k\neq i)}$ for $1\leq k\leq n+1$. Under our
assumptions, we have
$$\Pr(B)\mathbb{\leq\delta}_{n,p_{n}%
}+(n+1)\mathbb{\delta}_{n+1,1/n+1}$$.

\bigskip

\noindent We want to show that with high probability there exist constants
$c_{i}$ such that for all $i\in\{1,\ldots,n\}$, for all $z\in\mathcal{Z},$
$|f(Z_{1},\ldots,Z_{i},\ldots,Z_{n})-f(Z_{1},\ldots,Z_{i-1},z,Z_{i+1}%
,\ldots,Z_{n})|\leq c_{i}$.

\noindent Notice that%

\begin{align*}
|f(Z_{1},\ldots,Z_{i},\ldots,Z_{n})-f(Z_{1},\ldots,z,\ldots,Z_{n})|  &
=|(\mathbb{E}_{V_{n}^{tr}}\mathbb{P}_{n,V_{n}^{ts}}\psi_{V_{n}^{tr}%
}-\mathbb{P}\psi_{e_{n+1}^{n+1}})\\
& \quad -(\mathbb{E}_{V_{n}^{tr}}\mathbb{P}%
_{n,V_{n}^{ts}}^{^{\prime}}\psi_{V_{n}^{tr}}^{^{\prime}}-\mathbb{P}%
\psi_{e_{n+1}^{i}})|\\
&  \leq|\mathbb{E}_{V_{n}^{tr}}\mathbb{P}_{n,V_{n}^{ts}}\psi_{V_{n}^{tr}%
}-\mathbb{E}_{V_{n}^{tr}}\mathbb{P}_{n,V_{n}^{ts}}^{^{\prime}}\psi_{V_{n}%
^{tr}}^{^{\prime}}| \\
& \quad +|\mathbb{P}\psi_{e_{n+1}^{n+1}}-\mathbb{P}\psi
_{e_{n+1}^{i}}|.
\end{align*}

\noindent with $\mathbb{P}_{n,V_{n}^{tr}}^{^{\prime}}$ the weighted empirical
measure on the sample
$$\mathcal{E}_{n}=\{Z_{1},\ldots,Z_{i-1},z,Z_{i+1}%
,\ldots,Z_{n}\}$$
and $\psi_{V_{n}^{tr}}^{^{\prime}}$
the predictor trained on
$\mathcal{E}_{V_{n}^{tr}}$.

\bigskip

\noindent So, first, let us bound the second term, recall that
\[
|\mathbb{P(}\psi_{e_{n+1}^{n+1}}-\psi_{e_{n+1}^{i}})|\leq d(\psi
_{e_{n+1}^{n+1}},\psi_{e_{n+1}^{i}})\leq d(\psi_{e_{n+1}^{n+1}},\psi
_{n+1})+d(\psi_{n+1},\psi_{e_{n+1}^{i}})
\]
\noindent with $\psi_{n+1}$ trained on $\mathcal{D}_{n+1}=\{Z_{1}%
,\ldots,Z_{i-1},Z_{i},Z_{i+1},\ldots,Z_{n},z\}$. Thus, we have on $B^{\subset
}$, $|\mathbb{P}\psi_{e_{n+1}^{n+1}}-\mathbb{P}\psi_{e_{n+1}^{i}}|\leq
2(\frac{2\lambda}{n+1})^{\alpha}$.

\bigskip

\noindent To upper bound the first term, notice that%

\begin{align*}
|\mathbb{E}_{V_{n}^{tr}}\mathbb{P}_{n,V_{n}^{ts}}\psi_{V_{n}^{tr}}%
-\mathbb{E}_{V_{n}^{tr}}\mathbb{P}_{n,V_{n}^{ts}}^{^{\prime}}\psi_{V_{n}^{tr}%
}^{^{\prime}}|=  &  |\mathbb{E}_{V_{n}^{tr}}(\mathbb{P}_{n,V_{n}^{ts}}%
(\psi_{V_{n}^{tr}}-\psi_{V_{n}^{tr}}^{^{\prime}})|V_{n,i}^{tr}=1)\times
(1-p_{n})\\
&  \quad+\mathbb{E}_{V_{n}^{tr}}((\mathbb{P}_{n,V_{n}^{ts}}-\mathbb{P}%
_{n,V_{n}^{ts}}^{^{\prime}})\psi_{V_{n}^{tr}}|V_{n,i}^{ts}=1)\times p_{n}|.
\end{align*}

\noindent We always have for any $\psi$, $|(\mathbb{P}_{n,V_{n}^{ts}%
}-\mathbb{P}_{n,V_{n}^{ts}}^{^{\prime}})\psi|\leq1/np_{n}$ thus
$$|\mathbb{E}%
_{V_{n}^{tr}}((\mathbb{P}_{n,V_{n}^{ts}}-\mathbb{P}_{n,V_{n}^{ts}}^{^{\prime}%
})\psi_{V_{n}^{tr}}|V_{n,i}^{ts}=1)\times p_{n}|\leq1/n$$.

\bigskip

\noindent Until now, the previous lines hold independently of $d\in
\{d_{e},d_{1},d_{\infty}\}$.\ We still have to bound $|\mathbb{E}_{V_{n}^{tr}%
}(\mathbb{P}_{n,V_{n}^{ts}}(\psi_{V_{n}^{tr}}-\psi_{V_{n}^{tr}}^{^{\prime}%
})|V_{n,i}^{tr}=1)|$. In the particular case of the most stable kind of
stability (i.e. when $d=d_{\infty}$), we have
\[
|\mathbb{E}_{V_{n}^{tr}}(\mathbb{P}_{n,V_{n}^{ts}}(\psi_{V_{n}^{tr}}%
-\psi_{V_{n}^{tr}}^{^{\prime}})|V_{n,i}^{tr}=1)|\leq\mathbb{E}_{V_{n}^{tr}%
}(d_{\infty}(\psi_{V_{n}^{tr}},\psi_{V_{n}^{tr}}^{^{\prime}})|V_{n}^{tr}=1).
\]

\noindent On $B^{\subset}$, we get $d_{\infty}(\psi_{V_{n}^{tr}},\psi
_{V_{n}^{tr}}^{^{\prime}})\leq d_{\infty}(\psi_{V_{n}^{tr}},\psi
_{n+1})+d_{\infty}(\psi_{n+1},\psi_{V_{n}^{tr}}^{^{\prime}})\leq2(2\lambda
p_{n})^{\alpha}$.

\noindent Thus, on $B^{\subset}$, we have
\[
\mathbb{E}_{V_{n}^{tr}}(d_{\infty}(\psi_{V_{n}^{tr}},\psi_{V_{n}^{tr}%
}^{^{\prime}})|V_{n,i}^{tr}=1)\leq2(2\lambda p_{n})^{\alpha}.
\]

\noindent Putting all together, with probability at least $1-\delta_{n,p_{n}%
}^{^{\prime}}$,
\[
\sup_{1\leq i\leq n,z\in\mathcal{Z}}|f(Z_{1},\ldots,Z_{i},\ldots
,Z_{n})-f(Z_{1},\ldots,z,\ldots,Z_{n})|\leq5(2\lambda p_{n})^{\alpha}.
\]

\item $\widehat{R}_{CV}-\widetilde{R}_{n}$\textbf{\ is closed to zero with
high probability}

\bigskip

\noindent Applying theorem \ref{ch2:Kutin strong}, we obtain that
for all $\varepsilon \geq0$
\begin{align*}
\Pr(\widehat{R}_{CV}-\widetilde{R}_{n}\geq\varepsilon+\delta+\lambda
(2p_{n})^{\alpha})
&  \leq\Pr\mathbb{(}\widehat{R}_{CV}-\widetilde{R}_{n}-\mathbb{E}%
_{\mathcal{D}_{n}}(\widehat{R}_{CV}-\widetilde{R}_{n})\geq\varepsilon)\\
&  \leq2(\exp(-\frac{\varepsilon^{2}}{8n(5(2\lambda p_{n})^{\alpha}%
+\alpha^{^{\prime}})^{2}})+\frac{n}{\alpha^{^{\prime}}}\delta^{^{\prime}})\\
&  \leq2(\exp(-\frac{\varepsilon^{2}}{8(10\lambda)^{2}n(2\lambda
p_{n})^{2\alpha}})+\frac{n}{5(2\lambda p_{n})^{\alpha}}\delta^{^{\prime}%
}) \\
& \text{ by taking }\alpha^{^{\prime}}=5(2\lambda p_{n})^{\alpha}\text{.}%
\end{align*}
\end{enumerate}

\noindent By symmetry, we also have $\Pr\mathbb{(}\widehat{R}_{CV}%
-\widetilde{R}_{n}\leq-(\varepsilon+\delta_{n,p_{n}}+2\lambda p_{n}%
))\leq2(\exp(-\frac{\tau^{2}}{8(10\lambda)^{2}n(2\lambda p_{n})^{2\alpha}%
})+\frac{n}{5(2\lambda p_{n})^{\alpha}}\delta_{n,pn}^{^{\prime}})$ which
allows to conclude.

$\Box$

\begin{theorem}
[Strong stability]

Suppose that $\mathcal{H}$ holds. Let $\Psi$ be a machine
learning which is strong $(\lambda,(\delta_{n,p_{n}})_{n,p_{n}},d)$ stable.
Then, for all $\varepsilon\geq0$, we get%

\[
\Pr\mathbb{(}|\widehat{R}_{CV}-\widetilde{R}_{n}|\geq\varepsilon
+\lambda(2p_{n})^{\alpha})\leq2\exp(-2np_{n}\varepsilon^{2})+\kappa
(n)\delta_{n},
\]

\noindent where $\kappa(n)$ is the number of training vectors in the cross-validation.

\noindent Furthermore, if the distance $d$ is the uniform distance $d_{\infty
}$, then we have for any $\varepsilon\geq0$:%

\[
\Pr\mathbb{(}|\widehat{R}_{CV}-\widetilde{R}_{n}|\geq\varepsilon
+\delta_{n,p_{n}}+\lambda(2p_{n})^{\alpha})\leq4(\exp(-\frac{\varepsilon^{2}%
}{8n(5(2\lambda
p_{n})^{\alpha}+\alpha^{^{\prime}})^{2}})+\frac{n}{\alpha'
}\kappa(n)\delta_{n,p_{n}}^{^{\prime}}),
\]

\noindent with $\delta_{n,p_{n}}^{^{\prime}}=\delta_{n,p_{n}}%
+(n+1)\delta_{n,1/n}$. Thus, if we take $\alpha=5(2\lambda p_{n})^{\alpha}$,
we get
\[
\Pr\mathbb{(}|\widehat{R}_{CV}-\widetilde{R}_{n}|\geq\varepsilon
+\delta_{n,p_{n}}+\lambda(2p_{n})^{\alpha})\leq4(\exp(-\frac{\varepsilon^{2}%
}{8(10\lambda)^{2}n(2\lambda p_{n})^{2\alpha}})+\frac{n}{5(2\lambda
p_{n})^{\alpha}}\kappa(n)\delta_{n,p_{n}}^{^{\prime}}).
\]
\end{theorem}

\bigskip

\noindent{\bfseries Proof}

\noindent For the first inequality, it is sufficient to use
remarks \ref{Ch2:Rem1}.

\noindent For the second one, we can follow the previous proof, using remarks
\ref{Ch2:Rem1} and noticing that if we denote $B_{_{\mathbf{v}_{n}^{tr}}%
}:=\{d(\psi_{\mathbf{v}_{n}^{tr}},\psi_{n})\geq\lambda||\mathbb{P}%
_{n,\mathbf{v}_{n}^{tr}}-\mathbb{P}_{n}||\}$,\ then, we have%
\begin{align*}
\mathbb{P}^{\otimes n}d(\psi_{\mathbb{P}_{\mathbf{v}_{n}^{tr}}},\psi_{n})  &
=\mathbb{P}^{\otimes\mathbf{n}}(d(\psi_{\mathbf{v}_{n}^{tr}},\psi
_{n})|B_{\mathbf{v}_{n}^{tr}})\mathbb{P}^{\otimes\mathbf{n}}(B_{\mathbf{v}%
_{n}^{tr}})+\mathbb{P}^{\otimes\mathbf{n}}(d(\psi_{\mathbf{v}_{n}^{tr}}%
,\psi_{n})|B_{\mathbf{v}_{n}^{tr}}^{\subset})(1-\mathbb{P}^{\otimes\mathbf{n}%
}(B_{\mathbf{v}_{n}^{tr}}))\\
&  \leq1\times\delta_{n,p_{n}}+\lambda\mathbb{P}^{\otimes\mathbf{n}%
}||\mathbb{P}_{n,\mathbf{v}_{n}^{tr}}-\mathbb{P}_{n}||_{\alpha}\times
(1-\delta_{n,p_{n}})=\delta_{n,p_{n}}+\lambda(2p_{n})^{\alpha}(1-\delta
_{n,p_{n}}).
\end{align*}

\noindent Eventually, we get $\mathbb{P}^{\otimes n}(\widehat{R}%
_{CV}-\widetilde{R}_{n})\leq\delta_{n,p_{n}}+\lambda(2p_{n})^{\alpha}.$

$\Box$

\bigskip

\noindent Now, we derive results for the hold-out cross-validation which does
not make a symmetrical use of the dataset. We obtain

\begin{theorem}
[Strong stability and hold-out]\label{th:Hold-out}
\bigskip
Let $\Psi$ be a machine
learning which is strong $(\lambda,(\delta_{n,p_{n}})_{n,p_{n}},\delta)$
stable. Then the hold-out (or split sample) cross-validation satisfies for all
$\varepsilon\geq0$,%

\[
\Pr\mathbb{(}|\widehat{R}_{CV}-\widetilde{R}_{n}|\geq\varepsilon
+\lambda(2p_{n})^{\alpha})\leq2\exp(-2np_{n}\varepsilon^{2})+\delta_{n,p_{n}}.%
\]

\noindent Furthermore, if the distance is the uniform distance $d_{\infty}$,
then we have%

\begin{align*}
\Pr\mathbb(|\widehat{R}_{CV}-\widetilde{R}_{n}|\geq\varepsilon
+\delta_{n,p_{n}}+\lambda(2p_{n})^{\alpha}) &\leq4(\exp(-\frac{\varepsilon^{2}%
}{8(4\lambda(2p_{n})^{\alpha}+1/np_{n})^{2}}) \\
& \quad +\frac{n^{2}}{4\lambda
(2p_{n})^{\alpha}+1/np_{n}}\delta_{n,p_{n}}^{^{\prime}}),
\end{align*}

\noindent with $\delta_{n,p_{n}}^{^{\prime}}=\delta_{n,p_{n}}+n\delta_{n,1/n}$
\end{theorem}

\noindent{\bfseries Proof}

\noindent For the first inequality, it is enough to use remarks
\ref{Ch2:Rem1}.

\noindent For the second one, we start as previously. First, we bound in the
same way the expectation.

\bigskip

\noindent Secondly, we show that
$\widehat{R}_{CV}-\widetilde{R}_{n}$ is difference-bounded with
high probability.

\noindent Denote $f(Z_{1},Z_{2},\ldots,Z_{n}):=\widehat{R}_{CV}-\widetilde
{R}_{n}$. Let $z$ $\in$ $\mathcal{Z}$. Now denote as previously $B:=B_{1}\cup
B_{2}$ with $B_{1}=\{\frac{d(\psi_{\mathbf{v}_{n}^{tr}},\psi_{n}%
)}{||\mathbb{P}_{n,\mathbf{v}_{n}^{tr}}-\mathbb{P}_{n}||_{\alpha}}\geq
\lambda\}$ and $B_{2}=\{\sup_{i}\frac{d(\psi_{e_{n+1}^{i}},\psi_{n+1}%
)}{||\mathbb{P}_{n+1,e_{n+1}^{i}}-\mathbb{P}_{n+1}||_{\alpha}}\geq\lambda\}$.
Eventually, we have $\Pr(B)\leq\delta_{n,1-p_{n}}+n\delta_{n,1/n}$

\bigskip

\noindent We want to show that with high probability there exists constants
$c_{i}$ such that for all $i$, for all $z\in\mathcal{Z},$ $|f(Z_{1}%
,\ldots,Z_{i},\ldots,Z_{n})-f(Z_{1},\ldots,z,\ldots,Z_{n})|\leq c_{i}$. Since
$V_{n}^{tr}=\mathbf{v}_{n}^{ts}$ fixed vector in the case of hold-out, notice that:%

\begin{align*}
|f(Z_{1},\ldots,Z_{i},\ldots,Z_{n})-f(Z_{1},\ldots,z,\ldots,Z_{n})|  &
=|\mathbb{P}_{n,\mathbf{v}_{n}^{ts}}\psi_{\mathbf{v}_{n}^{tr}}-\mathbb{P}%
\psi_{e_{n+1}^{n+1}})-(\mathbb{P}_{n,\mathbf{v}_{n}^{ts}}^{^{\prime}}%
\psi_{\mathbf{v}_{n}^{tr}}^{^{\prime}}-\mathbb{P}\psi_{e_{n+1}^{i}})|\\
&  \leq|\mathbb{P}_{n,\mathbf{v}_{n}^{ts}}\psi_{\mathbf{v}_{n}^{tr}%
}-\mathbb{E}_{\mathbf{v}_{n}^{tr}}\mathbb{P}_{n,\mathbf{v}_{n}^{ts}}%
^{^{\prime}}\psi_{\mathbf{v}_{n}^{tr}}^{^{\prime}}|\\
& \quad  +|\mathbb{P}\psi
_{e_{n+1}^{n+1}}-\mathbb{P}\psi_{e_{n+1}^{i}}|,
\end{align*}

\noindent with $\mathbb{P}_{n,\mathbf{v}_{n}^{tr}}^{^{\prime}}$ the weighted
empirical measures of the sample $\mathcal{E}_{n}=\{Z_{1},\ldots
,Z_{i-1},z,Z_{i+1},\ldots,Z_{n}\}$ and $\psi_{\mathbf{v}_{n}^{tr}}^{^{\prime}%
}$ the predictor trained on $\mathcal{E}_{\mathbf{v}_{n}^{tr}}.$

\bigskip

\noindent So, first, let us bound the second term, recall that:

$$|\mathbb{P(}%
\psi_{e_{n+1}^{n+1}}-\psi_{e_{n+1}^{i}})|\leq d(\psi_{e_{n+1}^{n+1}}%
,\psi_{e_{n+1}^{i}})\leq d(\psi_{e_{n+1}^{n+1}},\psi_{n+1})+d(\psi_{n+1}%
,\psi_{e_{n+1}^{i}})$$.

\noindent Thus, on $B^{\subset}$, $|\mathbb{P}\psi_{e_{n+1}%
^{n+1}}-\mathbb{P}\psi_{e_{n+1}^{i}}|\leq2\lambda(\frac{2}{n+1})^{\alpha}.$

\bigskip

\noindent To upper bound the first term, notice that:%

\[
|\mathbb{P}_{n,\mathbf{v}_{n}^{ts}}\psi_{\mathbf{v}_{n}^{tr}}-\mathbb{P}%
_{n,\mathbf{v}_{n}^{ts}}^{^{\prime}}\psi_{\mathbf{v}_{n}^{tr}}^{^{\prime}%
}|=|\mathbb{P}_{n,\mathbf{v}_{n}^{ts}}(\psi_{\mathbf{v}_{n}^{tr}}%
-\psi_{\mathbf{v}_{n}^{tr}}^{^{\prime}})1_{\{\mathbf{v}_{n,i}^{tr}%
=1\}}+(\mathbb{P}_{n,\mathbf{v}_{n}^{ts}}-\mathbb{P}_{n,\mathbf{v}_{n}^{ts}%
}^{^{\prime}})\psi_{\mathbf{v}_{n}^{tr}}1_{\{\mathbf{v}_{n,i}^{ts}=1\}}|
\]

\noindent We always have for any $\psi$, $|(\mathbb{P}_{n,\mathbf{v}_{n}^{ts}%
}-\mathbb{P}_{n,\mathbf{v}_{n}^{ts}}^{^{\prime}})\psi|\leq1/np_{n}$ thus
$$|(\mathbb{P}_{n,\mathbf{v}_{n}^{ts}}-\mathbb{P}_{n,\mathbf{v}_{n}^{ts}%
}^{^{\prime}})\psi_{\mathbf{v}_{n}^{tr}}1_{\{\mathbf{v}_{n,i}^{ts}=1\}}%
|\leq1/np_{n}$$.

\bigskip

\noindent We still have to bound $|\mathbb{P}_{n,\mathbf{v}_{n}^{ts}}%
(\psi_{\mathbf{v}_{n}^{tr}}-\psi_{\mathbf{v}_{n}^{tr}}^{^{\prime}%
})1_{\{\mathbf{v}_{n,i}^{tr}=1\}}|$. As in the previous proof, we have when
$d=d_{\infty}$, $|\mathbb{P}_{n,\mathbf{v}_{n}^{ts}}(\psi_{\mathbf{v}_{n}%
^{tr}}-\psi_{\mathbf{v}_{n}^{tr}}^{^{\prime}})1_{\{\mathbf{v}_{n,i}^{tr}%
=1\}}|\leq d_{\infty}(\psi_{\mathbf{v}_{n}^{tr}},\psi_{\mathbf{v}_{n}^{tr}%
}^{^{\prime}})1_{\{\mathbf{v}_{n,i}^{tr}=1\}}$

\noindent On $B^{\subset}$, $d_{\infty}(\psi_{\mathbf{v}_{n}^{tr}}%
,\psi_{\mathbf{v}_{n}^{tr}}^{^{\prime}})\leq d_{\infty}(\psi_{\mathbf{v}%
_{n}^{tr}},\psi_{n+1})+d_{\infty}(\psi_{n+1},\psi_{\mathbf{v}_{n}^{tr}%
}^{^{\prime}})\leq2\lambda(2p_{n})^{\alpha}$. Thus, on $B^{\subset}$ we get
$d_{\infty}(\psi_{\mathbf{v}_{n}^{tr}},\psi_{\mathbf{v}_{n}^{tr}}^{^{\prime}%
})1_{\{\mathbf{v}_{n,i}^{tr}=1\}}\leq2\lambda(2p_{n})^{\alpha}$.

\bigskip

\noindent Putting all together, with probability at least $1-\delta_{n,p_{n}%
}^{^{\prime}}$,
\begin{align*}
\sup_{i,z}|f(Z_{1},\ldots,Z_{i},\ldots,Z_{n})-f(Z_{1},\ldots,z,\ldots
,Z_{n})| &\leq2\lambda(\frac{2}{n+1})^{\alpha}+\max((np_{n})^{-1},2\lambda
(2p_{n})^{\alpha})\\
& \leq4\lambda(2p_{n})^{\alpha}+(np_{n})^{-1}%
\end{align*}

\noindent To conclude, apply again theorem \ref{ch2:Kutin strong}.

$\Box$

\subsection{Weak stability}

\noindent We now derive results that stands for general
cross-validation procedures and weakly stable predictors. We
recall here the interest of the notion of weak stability. For
some class of machine learning, the notion of strong stability
may be too demanding. That is why weak stability is introduced.\
As a motivation, algorithms such as Adaboost satisfies the
following definition of weak stability.

\bigskip

\noindent We will use the definition of weak difference bounded introduced by
\cite{KUT02} and a corollary of his main theorem.

\begin{definition}
[Kutin\cite{KUT02}]Let $\Omega_{1},\ldots,\Omega_{n}$ be probability spaces.
Let $\Omega=\prod_{k=1}^{n}\Omega_{k}$ and let $X$ a random variable on
$\Omega$. We say that $X$ is weakly difference bounded by $(b,c,\delta)$ if
the following holds: for any $k$,
\[
\forall^{\delta}(\omega,v)\in\Omega\times\Omega_{k},\text{\ }\mathbb{P}%
(|X(\omega)-X(\omega^{^{\prime}})|)\leq c
\]
\noindent where $\omega_{k}^{^{\prime}}=v$ and $\omega_{i}^{^{\prime}}%
=\omega_{i}$ for $i\neq k$. and the notation $\forall^{\delta}\omega
,\Phi(\omega)$ means ''$\Phi(\omega)$ holds for all but but a $\delta$
fraction of $\Omega$''%

\[
|X(\omega)-X(\omega^{\prime})|\leq c
\]

\noindent Furthermore, for any $\omega,\omega^{\prime}\in\Omega$, differing
only one coordinate:
\[
|X(\omega)-X(\omega^{\prime})|\leq b
\]
\end{definition}

\noindent We will need the following theorem. It says in substance that a
weakly difference bounded function of independent variables is closed to its
expectation with probability.

\begin{theorem}
[Kutin\cite{KUT02}]\label{ch2:Kutin weak}Let
$\Omega_{1},\ldots,\Omega_{n}$ be probability spaces. Let
$\Omega=\prod_{k=1}^{n}\Omega_{k}$ and let $X$ a random variable
on $\Omega$.which is weakly difference bounded by $(b,c,\delta)$.
Assume $b\geq c\geq0$ and $\alpha>0$. Let $\mu=\mathbb{E}(X)$.
Then, for any
$\varepsilon>0$%

\[
\Pr(|X-\mu|\geq\varepsilon)\leq2\exp(-\frac{\varepsilon^{2}}{10nc^{2}%
(1+\frac{2\varepsilon}{15nc})^{2}})+\frac{2nb\delta^{1/2}}{c}\exp
(\frac{\varepsilon b}{4nc^{2}}))+2n\delta^{1/2}.%
\]
\end{theorem}

\begin{theorem}
[Cross-validation Weak stability]Suppose that $\mathcal{H}$ holds. Let $\Psi$
be a machine learning which is weak\textbf{\ }$(\lambda,(\delta_{n}%
)_{n},d,\mathbb{Q})$ stable with respect to the distance $d$. Then, for all
$\varepsilon\geq0$,%

\[
\Pr\mathbb{(}|\widehat{R}_{CV}-\widetilde{R}_{n}|\geq\varepsilon
+\lambda(2p_{n})^{\alpha})\leq2\exp(-2np_{n}\varepsilon^{2})+\delta_{n,p_{n}}%
\]

\noindent Furthermore, if the distance is the uniform distance $d_{\infty}$,
we have for all $\varepsilon\geq0$:%

\begin{align*}
\Pr\mathbb{(}|\widehat{R}_{CV}-\widetilde{R}_{n}|\geq\varepsilon
+\delta_{n,p_{n}}+\lambda(2p_{n})^{\alpha})&\leq4(\exp(-\frac{\varepsilon^{2}%
}{10n(5\lambda(2p_{n})^{\alpha})^{2}(1+\frac{2\varepsilon}{15n(5\lambda
(2p_{n})^{\alpha})})^{2}}) \\
& +\frac{2n\delta_{n,p_{n}}^{^{\prime}1/2}}%
{5\lambda(2p_{n})^{\alpha}}\exp(\frac{\varepsilon n}{4n(5\lambda
(2p_{n})^{\alpha})^{2}}))+n\delta_{n,p_{n}}^{^{\prime}1/2})),
\end{align*}

\noindent with $\delta_{n,p_{n}}^{\prime}=2\delta_{n,1/n}+\delta_{n,p_{n}}$
\end{theorem}

\noindent{\bfseries Proof}

\noindent In the following, denote $B$ the bad subset, i.e. $B=\cup
_{v_{n}^{tr}}B_{v_{n}^{tr}}$ with $B_{v_{n}^{tr}}=\{d(\psi_{\mathbb{P}%
_{n,v_{n}^{tr}}},\psi_{\mathbb{P}_{n}})\geq\lambda||\mathbb{P}_{n,v_{n}^{tr}%
}-\mathbb{P}_{n}||\}$. Since $\Psi$ is strong $(\lambda,(\delta_{n}%
)_{n,p_n},d,\mathbb{Q})$ stable, we have
$\mathbb{P(}B)\leq\delta_{n,p_{n}}$.

\bigskip

\begin{enumerate}
\item  For the general case, it is again sufficient to split $|\widehat
{R}_{CV}-\widetilde{R}_{n}|$ according to the same benchmark,
namely $\bar
{R}_{n(1-p)}=\mathbb{E}_{V_{n}^{tr}}\mathbb{P}\psi_{V_{n}^{tr}}$.

\noindent Thus,
\[
\Pr\mathbb{(}|\widehat{R}_{CV}-\widetilde{R}_{n}|\geq\varepsilon
+\lambda(2p_{n})^{\alpha})\leq\Pr\mathbb{(}|\widehat{R}_{CV}-\bar{R}%
_{n(1-p)}|\geq\varepsilon)+\Pr\mathbb{(}|\bar{R}_{n(1-p)}-\widetilde{R}%
_{n}|\geq\lambda(2p_{n})^{\alpha})
\]

\noindent The first term can be bounded as previously by $2\exp(-2np_{n}%
\varepsilon^{2})$.

\bigskip

\noindent For the second term, notice that $|\bar{R}_{n(1-p)}-\widetilde
{R}_{n}|=|\mathbb{E}_{V_{n}^{tr}}\mathbb{P}\psi_{V_{n}^{tr}}-\mathbb{E}%
_{V_{n}^{tr}}\mathbb{P}\psi_{n}|\leq\mathbb{E}_{V_{n}^{tr}}|\mathbb{P}%
\psi_{V_{n}^{tr}}-\mathbb{P}\psi_{n}|$. Recall that
$|\mathbb{P}\psi _{V_{n}^{tr}}-\mathbb{P}\psi_{n}|\leq
d(\psi_{V_{n}^{tr}},\psi_{n})$ and
$||\mathbb{P}_{n,V_{n}^{tr}}-\mathbb{P}_{n}||^{\alpha}=\lambda(2
p_{n})^{\alpha}$. Thus, since
$\Psi$ is weak $(\lambda,(\delta_{n,p_{n}})_{n,p_n},d)$ stable, we have%
\begin{align*}
\Pr\mathbb{(}|\hat{R}_{n(1-p_n)}-\widetilde{R}_{n}| \geq\lambda(2
p_{n})^{\alpha}) &
\leq\Pr\mathbb{(E}_{v_{n}^{tr}}d(\psi_{v_{n}^{tr}},\psi_{n})\geq
\lambda(2p_{n})^{\alpha})\\
&  \leq\Pr\mathbb{(\cup}_{v_{n}^{tr}}\{d(\psi_{v_{n}^{tr}},\psi_{n}%
)\geq\lambda||\mathbb{P}_{n,v_{n}^{tr}}-\mathbb{P}_{n}||_{\alpha}\})\\
&  =\Pr\mathbb{(\cup}_{v_{n}^{tr}}B_{v_{n}^{tr}})\leq\kappa(n)\delta_{n,p_{n}%
}\text{.}%
\end{align*}

\item  In the particular case, when $d=d_{\infty}$, we can also obtain a
stronger result.
\end{enumerate}

\noindent We proceed in three steps as in
\cite{BE02},\cite{KUNIY02} by using a bounded difference
inequality:

\begin{enumerate}
\item  first, we show that the expectation of $\widehat{R}_{CV}-\widetilde
{R}_{n}$ is small of the same order as for the strong stability.

\item  secondly, we show that the function $\widehat{R}_{CV}-\widetilde{R}%
_{n}$ seen as a function $f$ of $Z_{1},Z_{2},\ldots,Z_{n}$ is weakly
difference bounded, i.e. there exists constants $c_{1},\ldots,c_{n}$ such that
for all $i$, if $Z_{1},\ldots,Z_{i},\ldots,Z_{n},Z_{i^{\prime}}$ i.i.d. random
variables, we have with high probability

$$|f(Z_{1},\ldots,Z_{i},\ldots,Z_{n})-f(Z_{1},\ldots
,Z_{i^{\prime}},\ldots,Z_{n})|\leq c_{i}.$$

\item  finally, we use theorem \ref{ch2:Kutin weak} with the first two points to conclude.
\end{enumerate}

\bigskip

\begin{enumerate}
\item \textbf{The expectation of }$\widehat{R}_{CV}-\widetilde{R}_{n}%
$\textbf{\ is small}

\noindent As previously, denote $\mathbf{v}_{n}^{tr},\mathbf{v}_{n}^{ts}$
fixed vectors. We still have%

\[
\mathbb{P}^{\otimes n}(\widehat{R}_{CV}-\widetilde{R}_{n})=\mathbb{P}^{\otimes
n}(\mathbb{E}_{V_{n}^{tr}}\mathbb{P}_{n,V_{n}^{ts}}\psi_{V_{n}^{tr}%
}-\mathbb{P}\psi_{n})=\mathbb{P}^{\otimes n}\mathbb{P(}\psi_{\mathbf{v}%
_{n}^{tr}}-\psi_{n})\text{.}%
\]
\noindent since $\mathbb{P}^{\otimes n}\mathbb{E}_{V_{n}^{tr}}\mathbb{P}%
_{n,V_{n}^{ts}}\psi_{V_{n}^{tr}}=\mathbb{E}_{V_{n}^{tr}}\mathbb{P}^{\otimes
n}\mathbb{P}_{n,V_{n}^{ts}}\psi_{V_{n}^{tr}}=\mathbb{E}_{V_{n}^{tr}}%
\mathbb{P}^{\otimes n}\mathbb{P}\psi_{V_{n}^{tr}}=$ $\mathbb{P}^{\otimes
n}\mathbb{P}\psi_{\mathbf{v}_{n}^{tr}}$ where the first equality comes from
the linearity of expectation, the second from the fact that $\mathbb{P}%
_{n,V_{n}^{tr}}$ are independent of $\mathbb{P}_{n,V_{n}^{ts}}$,
and the third one from the $i.i.d.$ nature of $(Z_{i})_{i}.$

\bigskip

\noindent Recall that $\mathbb{P}(\psi_{\mathbf{v}_{n}^{tr}}-\psi_{n})\leq
d(\psi_{\mathbf{v}_{n}^{tr}},\psi_{n})$ where $d$ stands indifferently for
$d_{1},d_{e}$ or $d_{\infty}$. Thus, $\mathbb{P}^{\otimes n}\mathbb{P(}%
\psi_{\mathbf{v}_{n}^{tr}}-\psi_{n})\leq\mathbb{P}^{\otimes n}d(\psi
_{\mathbf{v}_{n}^{tr}},\psi_{n})$. By conditioning according to the small
values of $d(\psi_{\mathbf{v}_{n}^{tr}},\psi_{n})$, we obtain%

\begin{align*}
\mathbb{P}^{\otimes n}d(\psi_{\mathbf{v}_{n}^{tr}},\psi_{n})  &
=\mathbb{P}^{\otimes\mathbf{n}}(d(\psi_{\mathbf{v}_{n}^{tr}},\psi
_{n})|B_{\mathbf{v}_{n}^{tr}})\mathbb{P}^{\otimes\mathbf{n}}(B_{\mathbf{v}%
_{n}^{tr}}) \\
& \quad +\mathbb{P}^{\otimes\mathbf{n}}(d(\psi_{\mathbf{v}_{n}^{tr}}%
,\psi_{n})|B_{\mathbf{v}_{n}^{tr}}^{\subset})(1-\mathbb{P}^{\otimes\mathbf{n}%
}(B_{\mathbf{v}_{n}^{tr}}))\\
&  \leq1\times\delta_{n,p_{n}}+\lambda\mathbb{P}^{\otimes\mathbf{n}%
}||\mathbb{P}_{n,\mathbf{v}_{n}^{tr}}-\mathbb{P}_{n}||_{\alpha}\times
(1-\delta_{n,p_{n}})\leq\delta_{n,p_{n}}+\lambda(2p_{n})^{\alpha}\text{.}%
\end{align*}

\bigskip

\noindent Eventually, we still have $\mathbb{P}^{\otimes n}(\widehat{R}%
_{CV}-\widetilde{R}_{n})\leq\delta_{n,p_{n}}+\lambda(2p_{n})^{\alpha}$.

\item $\widehat{R}_{CV}-\widetilde{R}_{n}$\textbf{\ is difference bounded with
high probability}

\noindent Denote $f(Z_{1},Z_{2},\ldots,Z_{n}):=\widehat{R}_{CV}-\widetilde
{R}_{n}$.

\bigskip

\noindent We want to show that for all $i$, there exists constant $c_{i}$ such
$$|f(Z_{1},\ldots,Z_{i},\ldots,Z_{n})-f(Z_{1},\ldots,Z_{i}^{^{\prime}}%
,\ldots,Z_{n})|\leq c_{i}$$
with high probability where $Z_{1},\ldots
,Z_{i},\ldots,Z_{n},Z_{i}^{^{\prime}}$ are i.i.d. variables. Denote
$$B_{i}=\{\frac{d(\psi_{e_{n+1}^{i}},\psi_{n+1})}{||\mathbb{P}_{n+1,e_{n+1}%
^{i}}-\mathbb{P}_{n+1}||_{\alpha}}\geq\lambda\}.$$

\noindent We proceed as previously where
$(\cup_{v_{n}^{tr}}B_{v_{n}^{tr}})\cup B_{i}\cup B_{n+1}$ will
play the
role of $B$.%

\begin{align*}
|f(Z_{1},\ldots,Z_{i},\ldots,Z_{n})-f(Z_{1},\ldots,Z_{i}^{\prime},\ldots
,Z_{n})|  &
=|(\mathbb{E}_{V_{n}^{tr}}\mathbb{P}_{n,V_{n}^{ts}}\psi
_{V_{n}^{tr}}-\mathbb{P}\psi_{n})- \\
&\quad (\mathbb{E}_{V_{n}^{tr}}\mathbb{P}%
_{n,V_{n}^{ts}}^{^{\prime}}\psi_{V_{n}^{tr}}^{^{\prime}}-\mathbb{P}\psi_{_{n}%
}^{^{\prime}})|\\
&  \leq|\mathbb{E}_{V_{n}^{tr}}\mathbb{P}_{n,V_{n}^{ts}}\psi_{V_{n}^{tr}%
}-\mathbb{E}_{V_{n}^{tr}}\mathbb{P}_{n,V_{n}^{ts}}^{^{\prime}}\psi_{V_{n}%
^{tr}}^{^{\prime}}| \\
&\quad +|\mathbb{P}\psi_{n}-\mathbb{P}\psi_{n}^{^{\prime}}|,
\end{align*}

\noindent with $\mathbb{P}_{n}^{^{\prime}},\mathbb{P}_{n,V_{n}^{ts}}%
^{^{\prime}}$ the weighted empirical measures of the sample
$$\mathcal{D}%
_{n}^{^{\prime}}=\{Z_{1},\ldots,Z_{i}^{^{\prime}},\ldots,Z_{n}\}$$
 and
$\psi_{n}^{^{\prime}}$ the predictor built on $\mathcal{D}_{n}^{^{\prime}}$.

\bigskip

\noindent So, first, let us bound the second term, recall that: $|\mathbb{P}%
(\psi_{n}-\psi_{n}^{^{\prime}})|\leq d(\psi_{n},\psi_{n}^{^{\prime}})\leq
d(\psi_{n},\psi_{n+1})+d(\psi_{n+1},\psi_{n}^{^{\prime}})$.with $\psi_{n+1}$
the predictor trained on the sample $\mathcal{D}_{n+1}=\{Z_{1},\ldots
,Z_{i},\ldots,Z_{n},Z_{i^{\prime}}\}$. Thus, on $B^{\subset}$, we have $|\mathbb{P}\psi_{n}-\mathbb{P}\psi_{n}^{^{\prime}%
}|\leq2\lambda(2/n)^{\alpha}$.

\bigskip

\noindent To upper bound the first term, notice that%

\begin{align*}
{\small |E}_{V_{n}^{tr}}{\small P}_{n,V_{n}^{ts}}{\small \psi}_{V_{n}^{tr}%
}{\small -E}_{V_{n}^{tr}}{\small P}_{n,V_{n}^{ts}}^{^{\prime}}{\small \psi
}_{V_{n}^{tr}}^{^{\prime}}{\small |}  &  {\small =|E}_{V_{n}^{tr}}%
{\small (P}_{n,V_{n}^{ts}}{\small (\psi}_{V_{n}^{tr}}{\small -\psi}%
_{V_{n}^{tr}}^{^{\prime}}{\small )|V}_{n,i}^{tr}{\small =1)\times(1-p}%
_{n}{\small )}\\
&  {\small +E}_{V_{n}^{tr}}{\small ((P}_{n,V_{n}^{ts}}{\small -P}%
_{n,V_{n}^{ts}}^{^{\prime}}{\small )\psi}_{V_{n}^{tr}}{\small |V}_{n,i}%
^{ts}{\small =}{\small 1)\times p}_{n}{\small |}.%
\end{align*}

\noindent We always have for all $\psi$, $|(\mathbb{P}_{n,V_{n}^{ts}%
}-\mathbb{P}_{n,V_{n}^{ts}}^{^{\prime}})\psi|\leq1/np_{n}$ thus we get $$|\mathbb{E}%
_{V_{n}^{tr}}((\mathbb{P}_{n,V_{n}^{ts}}-\mathbb{P}_{n,V_{n}^{ts}}^{^{\prime}%
})\psi_{V_{n}^{tr}},V_{n}^{ts}=1)\times p_{n}|\leq1/n$$

\bigskip

\noindent We still have to bound
$$|\mathbb{E}_{V_{n}^{tr}}(\mathbb{P}%
_{n,V_{n}^{ts}}(\psi_{V_{n}^{tr}}-\psi_{V_{n}^{tr}}^{^{\prime}})|V_{n,i}%
^{tr}=1)|\leq\mathbb{E}_{V_{n}^{tr}}(d_{\infty}(\psi_{V_{n}^{tr}},\psi
_{V_{n}^{tr}}^{^{\prime}})|V_{n,i}^{tr}=1)$$

\noindent On $B_{v_{n}^{tr}}^{\subset}$, $d_{\infty}(\psi_{v_{n}^{tr}}%
,\psi_{v_{n}^{tr}})\leq
d_{\infty}(\psi_{v_{n}^{tr}},\psi_{n+1})+d_{\infty
}(\psi_{n+1},\psi_{v_{n}^{tr}}^{^{\prime}})\leq2\lambda(2p_{n})^{\alpha}$.

Thus, we get
$\mathbb{E}_{V_{n}^{tr}}(d_{\infty}(\psi_{V_{n}^{tr}},\psi
_{V_{n}^{tr}}^{^{\prime}}),V_{n}^{tr}=1)\leq2\lambda(2p_{n})^{\alpha}$
on $(\cup_{v_{n}^{tr}}B_{v_{n}^{tr}})^{\subset}.$

\bigskip

\noindent Putting all together, with probability at least $1-2\delta
_{n+1,1/(n+1)}-\delta_{n,p_{n}}$,
$$|f(Z_{1},\ldots,Z_{i},\ldots,Z_{n}%
)-f(Z_{1},\ldots,Z_{i^{\prime}},\ldots,Z_{n})|\leq5\lambda(2p_{n})^{\alpha}.$$

\item $\widehat{R}_{CV}-\widetilde{R}_{n}$\textbf{\ is closed to zero with
high probability}

\bigskip

\noindent Applying theorem \ref{ch2:Kutin weak}, we obtain for all
$\varepsilon \geq0$:
\begin{align*}
\Pr\mathbb{(}\widehat{R}_{CV}-\widetilde{R}_{n}\geq\varepsilon+\delta
_{n,p_{n}}+\lambda(2p_{n})^{\alpha})  &  \leq2(\exp(-\frac{\varepsilon^{2}%
}{10n(5\lambda(2p_{n})^{\alpha})^{2}(1+\frac{2\varepsilon}{15n(5\lambda
(2p_{n})^{\alpha})})^{2}}) \\
&\quad +\frac{2n\delta_{n,p_{n}}^{^{\prime}1/2}}%
{5\lambda(2p_{n})^{\alpha}}\exp(\frac{\varepsilon n}{4n(5\lambda
(2p_{n})^{\alpha})^{2}}))+n\delta_{n,p_{n}}^{^{\prime}1/2})\\
&  \leq2(\exp(-\frac{\varepsilon^{2}}{10n(5\lambda(2p_{n})^{\alpha}%
)^{2}(1+\frac{2\varepsilon}{15n(5\lambda(2p_{n})^{\alpha})})^{2}})\\
& \quad +\frac{2n\delta_{n,p_{n}}^{^{\prime}1/2}}{5\lambda(2p_{n})^{\alpha}}\exp
(\frac{\varepsilon n}{4n(5\lambda(2p_{n})^{\alpha})^{2}}))+n\delta_{n,p_{n}%
}^{^{\prime}1/2}).
\end{align*}

\bigskip
\end{enumerate}

\noindent By symmetry, we also upper bound $\Pr\mathbb{(}\widehat{R}%
_{CV}-\widetilde{R}_{n}\leq-(\varepsilon+\delta_{n,p_{n}}+\lambda
(2p_{n})^{\alpha}))$ by the same quantity.

$\Box$

\bigskip

\begin{theorem}
[Weak stability]Suppose that $\mathcal{H}$ holds. Let $\Psi$ be a machine
learning which is weak $(\lambda,(\delta_{n,p_{n}})_{n,p_{n}},d)$. Then for
all $\varepsilon\geq0$, we have%

\[
\Pr\mathbb{(}|\widehat{R}_{CV}-\widetilde{R}_{n}|\geq\varepsilon
+\lambda(2p_{n})^{\alpha})\leq2\exp(-2np_{n}\varepsilon^{2})+\kappa
(n)\delta_{n,p_{n}}%
\]

\noindent where $\kappa(n)$ is the number of elements in the cross-validation.

\noindent Furthermore, if the distance is the uniform distance $d_{\infty}$,
we have%

\begin{align*}
\Pr\mathbb{(}|\widehat{R}_{CV}-\widetilde{R}_{n}|\geq\varepsilon
+\delta_{n,p_{n}}+\lambda(2p_{n})^{\alpha})&\leq4(\exp(-\frac{\varepsilon^{2}%
}{10n(5\lambda(2p_{n})^{\alpha})^{2}(1+\frac{2\varepsilon}{15n(5\lambda
(2p_{n})^{\alpha})})^{2}}) \\
& \quad +\frac{2n\delta_{n,p_{n}}^{^{\prime}1/2}}%
{5\lambda(2p_{n})^{\alpha}}\exp(\frac{\varepsilon n}{4n(5\lambda
(2p_{n})^{\alpha})^{2}}))+n\delta_{n,p_{n}}^{^{\prime}1/2}))
\end{align*}

\noindent with $\delta_{n,p_{n}}^{^{\prime}}=\delta_{n,1/n}+\kappa
(n)\delta_{n,p_{n}}$
\end{theorem}

\bigskip

\noindent{\bfseries Proof.}

\bigskip

\noindent For the first inequality, it is enough to use remarks
\ref{Ch2:Rem1}.

\noindent For the second, it is enough to follow the previous
proofs and to notice that
$\Pr(\cup_{v_{n}^{tr}}B_{v_{n}^{tr}})\leq\kappa(n)\delta_{n,p_{n}}$.

$\Box$

\bigskip

\noindent Similar results for hold-out can be derived in the spirit of
proposition \ref{th:Hold-out}. We can now use the previous probability upper
bounds to derive upper bounds for the expectation of $|\widehat{R}%
_{CV}-\widetilde{R}_{n}|$.

\subsection{Results for the $L_{1}$ norm}

\noindent For the sake of simplicity, we suppose here that
$\alpha=1$. In the general case, we just consider the weakest
notion: weak $(\lambda ,(\delta_{n,p_{n}})_{n,p_{n}},d)$
stability.

\begin{theorem}
[$L_{1}$ norm of cross-validation estimate]Suppose that $\mathcal{H}$ holds.
Let $\Psi$ be a machine learning which is weak $(\lambda,(\delta_{n,p_{n}%
})_{n,p_{n}},d)$ stable. Then, we have%

\[
\mathbb{E}_{\mathcal{D}_{n}}|\widehat{R}_{CV}-\widetilde{R}_{n}|\leq2\lambda
p_{n}+\sqrt{\frac{2}{np_{n}}}+\delta_{n,p_{n}}.%
\]

\noindent Furthermore, if $\Psi$ is a machine learning which is strong
$(\lambda,(\delta_{n})_{n},d_{\infty},\mathbb{Q})$ stable, we have%

\[
\mathbb{E}_{\mathcal{D}_{n}}|\widehat{R}_{CV}-\widetilde{R}_{n}|\leq
\delta_{n,p_{n}}+2\lambda p_{n}+51\lambda\sqrt{n}p_{n}+\frac{n}{9\lambda
p_{n}}\delta_{n,p_{n}}^{^{\prime}},%
\]

\noindent with $\delta_{n,p_{n}}^{^{\prime}}=\delta_{n,p_{n}}+(n+1)\delta
_{n+1,1/n+1}$
\end{theorem}

{\bfseries Proof.}

\noindent These inequalities are a consequence of the previous
propositions and of the following lemma (for a proof, see e.g.
\cite{DGL96}):

\begin{lemma}
\label{lem esperance} Let $X$ be a nonnegative random variable. Let $K,C$
nonnegative real such that $C\geq1$. Suppose that for all $\varepsilon>0$,
$\mathbb{P}(X\geq\varepsilon)\leq C\exp(-K\varepsilon^{2})$. Then:
\[
\mathbb{E}X\leq\sqrt{\frac{\ln(C)+2}{K}}.
\]
\end{lemma}

\noindent For the second one, it is enough to follow the previous proofs and
to notice that $\Pr(\cup_{V_{n}^{tr}}B_{V_{n}^{tr}})\leq\kappa(n)\delta
_{n,p_{n}}$

$\Box$

\bigskip

\noindent We deduce that

\begin{corollary}
Suppose that $\mathcal{H}$ holds. If $\Psi$ be a machine learning which is
weak $(\lambda,(\delta_{n,p_{n}})_{n,p_{n}},d)$ stable, we define the
splitting rule $p_{n}^{\star}=(1/\sqrt{2}4\lambda)^{2/3}(1/n)^{1/3}$. Then, we have%

\[
\mathbb{E}_{\mathcal{D}_{n}}|\widehat{R}_{CV}-\widetilde{R}_{n}|\leq
4(\lambda/n)^{1/3}.
\]

\noindent Furthermore, if $\Psi$ is a machine learning which is strong
$(\lambda,(\delta_{n})_{n},d_{\infty},\mathbb{Q})$ stable, we use
leave-one-out cross-validation for $n$ large enough. And we have%

\[
\mathbb{E}_{\mathcal{D}_{n}}|\widehat{R}_{CV}-\widetilde{R}_{n}|=O_{n}%
(\lambda/\sqrt{n}).
\]
\end{corollary}

\noindent{\bfseries Proof.}

\bigskip

\noindent Recall that for a large class of learning algorithm, we
have in mind that
$\delta_{n,p_{n}}=O_{n}(p_{n}\exp(-n(1-p_{n}))$. Thus $2\lambda
p_{n}+\sqrt{\frac{2}{np_{n}}}+\delta_{n,p_{n}}\leq4\lambda
p_{n}+\sqrt {\frac{2}{np_{n}}}$. We can differentiate this last
bound seen as a function of $p_{n}$. We obtain
$p_{n}^{\star}=(1/\sqrt{2}4\lambda)^{2/3}(1/n)^{1/3}$.
Thus, we deduce that $\mathbb{E}_{\mathcal{D}_{n}}|\widehat{R}_{CV}%
-\widetilde{R}_{n}|\leq4(\lambda/n)^{1/3}$. If $\Psi$ is a machine learning
which is strong $(\lambda,(\delta_{n})_{n},d_{\infty},\mathbb{Q})$ stable, we
obtain $\delta_{n,p_{n}}+2\lambda p_{n}+51\lambda\sqrt{n}p_{n}+\frac
{n}{9\lambda p_{n}}\delta_{n,p_{n}}^{^{\prime}}\leq4\lambda p_{n}%
+51\lambda\sqrt{n}p_{n}$ for $n$ large enough since $\frac{n}{9\lambda p_{n}%
}\delta_{n,p_{n}}^{^{\prime}}=O_{n}(n^{3}\exp(-n/2))$ if
$p_{n}\leq1/2$. Thus,
$p_{n}^{\star}=1/n$ for $n$ large enough and $\mathbb{E}_{\mathcal{D}_{n}%
}|\widehat{R}_{CV}-\widetilde{R}_{n}|=O_{n}(\lambda/\sqrt{n})$.

\bigskip$\Box$

\noindent We have obtained the following conclusions:

\begin{itemize}
\item  Cross-validation is consistent as an estimator of the generalization
error of stable algorithms.

\item  There is a tradeoff interpretation in the choice of the proportion of
elements $p_{n}$ of the test set: the smaller $p_{n}$ is, the greater the term
$B(n,p_{n},\varepsilon)$ is controlled but the less the term $V(n,p_{n}%
,\varepsilon)$ is upper bounded.

\item  In the general setting, our bounds require that the sizes of the
training set and the test set grow to infinity.

\item  In the particular case of the stability with respect to the most stable
kind of stability (namely the uniform stability), we can have a stronger
result: the number of elements in the test set does need to grow to infinity
for the consistency of symmetric cross-validation procedures. But we lose this
property with the hold-out cross-validation.

\item  Symmetric cross-validation out performs hold-out cross-validation for
large sets.

\item  At last, as far as the expectation $\mathbb{E}_{\mathcal{D}_{n}%
}|\widehat{R}_{CV}-\widetilde{R}_{n}|$ is concerned, we can define a splitting
rule in the general setting.
\end{itemize}

\newpage

\section{Appendices}

\subsection{Inequalities}

\noindent We recall three very useful results. The first one, due to
\cite{HOEF63}, bounds the difference between the empirical mean and the
expected value. The second one, due to \cite{VC71}, bounds the supremum over
the class of predictors of the difference between the training error and the
generalization error. The last one is called the bounded differences
inequality \cite{McD89}.

\begin{theorem}
[Hoeffding's inequality]Let $X_{1,\ldots,}X_{n}$ independent
random variables in $[a_{i},b_{i}]$. Then for all
$\varepsilon>0,$ we get
\[
\mathbb{P}(\sum X_{i}-\mathbb{E}(\sum X_{i})\geq n\mathbb{\varepsilon})\leq
e^{-\frac{2\varepsilon^{2}}{\sum_{i}(b_{i-}a_{i})^{2}}}.%
\]
\end{theorem}

\begin{theorem}
[McDiarmid, \cite{McD89}]Let $X_{1,\ldots,}X_{n}$ be independent
random variables taking
values in a set $A$, and assume that $f:A^{n}\rightarrow\mathcal{R}$ satisfies%
\[
\forall i,\sup_{\substack{x_{1},\ldots,x_{i},\ldots,x_{n} \\x_{i}^{^{\prime}}%
}}|f(x_{1},...,x_{n})-f(x_{1},...,x_{i^{\prime}},...,x_{n})|\leq c_{i}.%
\]
Then for all $\varepsilon>0,$ we have%
\[
\mathbb{P(}f(X_{1},...,X_{n})-\mathbb{E}f(X_{1},...,X_{n})\geq\varepsilon)\leq
e^{-\frac{2\varepsilon^{2}}{\sum_{i}c_{i}^{2}}}.%
\]
\end{theorem}

\end{document}